\pdfoutput=1

\documentclass{article}
\usepackage[font=small,labelfont=bf]{caption} 
\usepackage{subcaption}                      
\usepackage{booktabs}    
\usepackage{multirow}    
\usepackage[normalem]{ulem} 
\usepackage{array} 
\usepackage[table]{xcolor}
\usepackage[]{EMNLP2023}
\usepackage{verbatim}
\usepackage{fancyvrb}
\usepackage{algorithm}
\usepackage{framed}
\usepackage{times}
\usepackage{latexsym}
\usepackage{amsmath,mathtools}
\usepackage{graphicx}
\usepackage{amsfonts}
\usepackage{float}
\usepackage{algpseudocode}

\usepackage[T1]{fontenc}

\usepackage[utf8]{inputenc}

\usepackage{microtype}

\usepackage{inconsolata}
\usepackage{siunitx}
\title{EvolKV: Evolutionary KV Cache Compression for LLM Inference}

\author{Bohan Yu$^{1,2}$ \and Yekun Chai$^3$\thanks{\,\,\,Corresponding author.} \\
         $^1$School of Advanced Interdisciplinary Sciences, \\ University of Chinese Academy of Sciences, Beijing, China \\
  $^2$The Key Laboratory of Cognition and Decision Intelligence for Complex Systems,\\ Institute of Automation, CAS, Beijing, China\\
  $^3$ETH Zurich \\ \texttt{yubohan23@mails.ucas.ac.cn} \quad \texttt{yechai@ethz.ch}}

\begin{document}
\maketitle
\begin{abstract}
Existing key-value (KV) cache compression methods typically rely on heuristics, such as uniform cache allocation across layers or static eviction policies, however, they ignore the critical interplays among layer-specific feature patterns and task performance, which can lead to degraded generalization. In this paper, we propose EvolKV, an adaptive framework for layer-wise, task-driven KV cache compression that jointly optimizes the memory efficiency and task performance. By reformulating cache allocation as a multi-objective optimization problem, EvolKV leverages evolutionary search to dynamically configure layer budgets while directly maximizing downstream performance. Extensive experiments on 11 tasks demonstrate that our approach outperforms all baseline methods across a wide range of KV cache budgets on long-context tasks and surpasses heuristic baselines by up to 7 percentage points on GSM8K. Notably, EvolKV achieves superior performance over the full KV cache setting on code completion while utilizing only 1.5\% of the original budget, suggesting the untapped potential in learned compression strategies for KV cache budget allocation. 

\end{abstract}

\section{Introduction}
Key-value (KV) cache~\cite{shi2024costdownreviewmethods,cai2024pyramidkvdynamickvcache,xiao2024efficientstreaminglanguagemodels,li2024snapkvllmknowslooking} has become a cornerstone of efficient inference in large language models (LLMs)~\cite{openai2024gpt4technicalreport,touvron2023llama2openfoundation,Multi-viewfusionforinstructionminingoflargelanguagemodel,bai2023qwentechnicalreport, lozhkov2024starcoder}, allowing them to reuse previously computed hidden states and thus reduce redundant computation. However, the memory footprint of a KV cache scales linearly with input sequence length, and the quadratic complexity of self-attention makes long-range inference prohibitively slow when the full cache is retained.

To address these challenges, existing KV cache compression methods predominantly rely on rule-based heuristics. Current approaches can be categorized into three paradigms: (1) fixed-position retention across all layers~\cite{child2019generatinglongsequencessparse,beltagy2020longformerlongdocumenttransformer,xiao2024efficientstreaminglanguagemodels}, (2) uniform layer allocation with attention-weighted eviction~\cite{li2024snapkvllmknowslooking,ge2024modeltellsdiscardadaptive,zhang2023h2oheavyhitteroracleefficient,liu2023scissorhandsexploitingpersistenceimportance,Transformers-are-Multi-State-RNNs}, and (3) pyramidal strategies with predefined depth-wise attenuation~\cite{cai2024pyramidkvdynamickvcache,yang2024pyramidinferpyramidkvcache}. While effectively for memory reduction, these heuristics fail to account for two critical aspects: (1) the varying functional roles of transformer layers in information processing~\cite{wang2020rethinkingvaluetransformercomponents,zhang2024investigatinglayerimportancelarge,skean2025layerlayeruncoveringhidden}, (2) the dynamic relationship between cache and task performance. Relying solely on rule-based allocation of KV cache budgets across layers can lead to suboptimal retention of task-relevant information. 

\begin{figure*}[htbp]
\centerline{\includegraphics[scale=0.615]{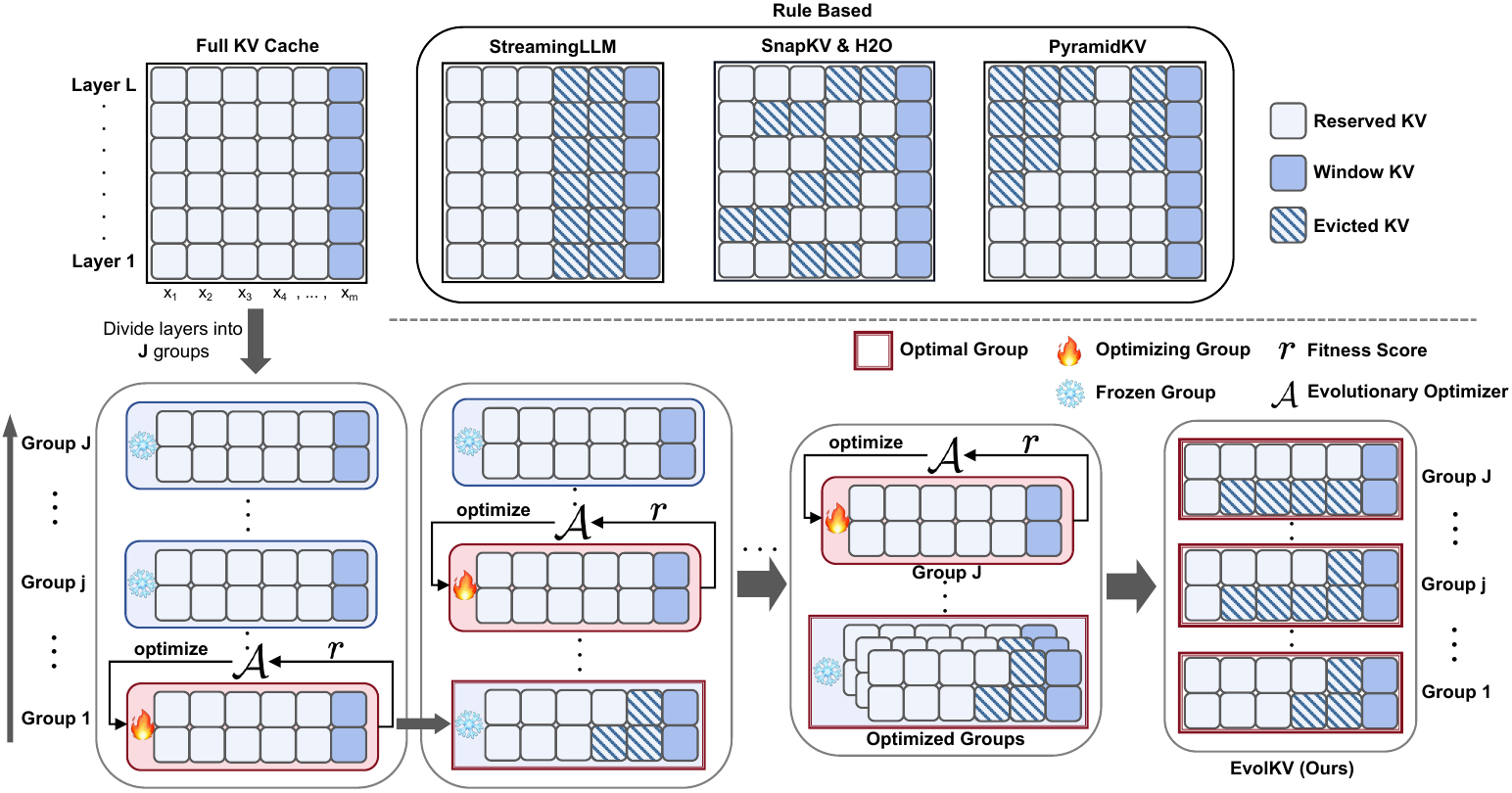}}
\caption{Illustration of the EvolKV framework. Compared to rule-based strategies (top row), EvolKV performs bottom-up, group-wise KV cache budget optimization using evolutionary search, progressively refining each layer group based on task-specific fitness feedback.}
\label{ov}
\end{figure*}

In response to these limitations, we employ evolutionary algorithms~\cite{10.7551/mitpress/Adaptation-in-Natural-and-Artificial-Systems,Differential-Evolution-A-Simple-and-Efficient-Heuristic-for-Global-Optimization-over-Continuous-Spaces} to directly search for optimal KV cache allocation based on the task performance, inspired by~\cite{chai-etal-2022-clip}. We introduce EvolKV, an evolutionary framework that adaptively allocates KV cache budgets across transformer layers, as shown in Figure~\ref{ov}. It formulates per-layer KV cache budgets as optimization variables, partitions them into groups, and employs an evolutionary algorithm~\cite{CMA-ES} to iteratively search for group-wise configurations that directly maximize downstream task fitness scores. By integrating task-driven optimization with layer-specific cache pruning, EvolKV achieves fine-grained, performance-aware allocation aligned with the varying contributions of different layers. 

In contrast to rigid heuristics, EvolKV provides a flexible and effective mechanism for layer-wise KV cache budget allocation guided by downstream task objectives. First, it formulates layer/group-wise cache budget as learnable parameters, in which we group layers into optimization units for efficient search. Then, we directly maximize the performance of downstream tasks using black-box evolutionary optimization methods. By doing so, our approach enables task-aware, granular cache allocation that automatically adapts to each group or layer's functional contribution. Specifically, it can accommodate diverse evaluation criteria, such as accuracy and F1 score, and discover non-uniform distributions (\textit{i.e.}, patterns deviated from heuristic fixed-length or pyramidal patterns) without predefined assumptions. 




We conduct comprehensive experiments on Mistral-7B-Instruct and Llama-3-8B-Instruct, evaluating EvolKV across four distinct benchmarks (eleven tasks), covering long-context retrieval, long-context reasoning, and mathematical tasks. Our results demonstrate that task-optimized KV cache allocation yields consistent improvements: (1) On the Needle-in-a-Haystack benchmark, EvolKV achieves up to a 13\% improvement over the best baseline. (2) In the RULER benchmark, EvolKV delivers up to a 3.6\% gain over the strongest baseline. (3) Across LongBench evaluations, it consistently outperforms all baseline methods across a wide range of target KV cache budgets (ranging from 128 to 2048), and remarkably exceeds the full model’s performance while utilizing only 1.5\% of its KV cache budget. (4) For GSM8K, EvolKV achieves up to a 7 percentage points improvement in accuracy over the strongest baseline under a 128 KV cache budget, preserving up to 95.7\% of the full-model performance, while the strongest baseline retains only up to 84.5\% under a 512 KV cache budget.

In conclusion, our key contributions are as follows:
\begin{itemize}
    \item We propose EvolKV, the first framework to formulate layer-wise KV cache budget allocation as a black-box optimization problem. 
    \item EvolKV operates on frozen LLMs, supports arbitrary evaluation metrics, without requiring fine-tuning or architectural modifications.
    \item Empirical results show that task-aware KV cache allocations consistently diverge from conventional heuristics, favoring non-uniform distributions that depart from fixed or pyramidal rules. EvolKV consistently outperforms strong baselines across long-context and reasoning tasks, even surpassing full-model performance under extreme compression.
\end{itemize}

\section{Related Work}\label{sec:related works}
\paragraph{KV Cache Compression}The substantial parameter counts of LLMs present significant challenges for inference. To overcome this difficulty, many efforts have been made to improve the inference efficiency of LLMs, such as the methods described in~\cite{zhang2023h2oheavyhitteroracleefficient,Sheng2023HighthroughputGI,liu2023scissorhandsexploitingpersistenceimportance,li2024snapkvllmknowslooking,Transformers-are-Multi-State-RNNs}, which evict the KV caches with the lowest accumulated attention weights. StreamingLLM~\cite{xiao2024efficientstreaminglanguagemodels} identifies the phenomenon of attention sink by retaining only the initial and most recent KV caches, thus maintaining a fixed positional alignment in the KV cache. These methods maintain a uniform KV cache budget across all layers, disregarding the actual demand for KV cache in each individual layer. Meanwhile, previous studies such as~\cite{cai2024pyramidkvdynamickvcache, yang2024pyramidinferpyramidkvcache} have reported a progressive reduction in the number of important KV cache across layers, forming a pyramid-shaped distribution. However, existing methods often set an attenuation coefficient to control the KV cache budget in each layer, thus ignoring that the cache budget actual needed in each layer do not necessarily exhibit a monotonically decreasing pattern. Instead of rule-based or heuristic methods, this study aims to compress the KV cache through an evolutionary, layer-wise optimization approach.

\paragraph{Evolutionary Algorithms}These are stochastic optimization algorithms inspired by the principles of natural evolution~\cite{10.7551/mitpress/Adaptation-in-Natural-and-Artificial-Systems,Optimierung-technischer-Systeme-nach-Prinzipien-der-biologischen-Evolution,alam2020geneticalgorithmreviewsimplementations}. It is widely used to address complex optimization and search problems by iteratively evolving high-quality solutions through the simulation of biological mechanisms such as selection, crossover (recombination), and mutation~\cite{Genetic-programming:-on-programming-computers-by-means-of-natural-selection,Particle-swarm-optimization,Differential-Evolution-A-Simple-and-Efficient-Heuristic-for-Global-Optimization-over-Continuous-Spaces,CMA-ES}. In this study, we leverage evolutionary algorithms in conjunction with downstream task performance to optimize the configuration of layer-wise KV cache budgets.

\section{Evolutionary KV Cache Compression}
\subsection{Motivation}
We observe that the existing methods mainly present the following categories in the KV cache budget allocation of layers:
\begin{itemize}
    \item \textbf{Fixed 
 Position:} Each layer preserves KV caches at the same position~\cite{child2019generatinglongsequencessparse,beltagy2020longformerlongdocumenttransformer,xiao2024efficientstreaminglanguagemodels}.
    \item \textbf{Identical Budget:} Each layer retains the same budget of KV caches, but their positions vary across layers~\cite{zhang2023h2oheavyhitteroracleefficient,liu2023scissorhandsexploitingpersistenceimportance,li2024snapkvllmknowslooking}.
    \item \textbf{Pyramidal Allocation:} The KV cache budget allocation follows a pyramidal pattern, decreasing progressively across layers~\cite{cai2024pyramidkvdynamickvcache,yang2024pyramidinferpyramidkvcache}.
\end{itemize}

Previous studies~\cite{wang2020rethinkingvaluetransformercomponents, zhang2024investigatinglayerimportancelarge, skean2025layerlayeruncoveringhidden} have shown that different layers of LLMs vary in their importance and process information at different levels of granularity. However, most existing compression methods overlook such heterogeneity and instead adopt rule-based or heuristic strategies for KV cache compression, which often results in suboptimal inference performance. These observations highlight the necessity of adapting the KV cache budget individually for each layer. 


\subsection{EvolKV}
To address the limitations of rule-based or heuristic allocation strategies, as shown in Figure~\ref{ov}, we introduce EvolKV, a dynamic and task-driven evolutionary framework that adaptively allocates the KV cache budget for each layer by leveraging performance feedback from downstream tasks. We present a comparison of budget allocations between EvolKV and other methods in Figure~\ref{fig:sub Layer_KV_Budget_comparison}.

\paragraph{Optimization Objectives of Evolutionary Compression}
Evolutionary algorithms generate candidate solutions and evaluate their fitness, iteratively refining the search strategy based on fitness feedback to progressively guide the population toward better solutions. In this paper, EvolKV treats performance feedback from downstream tasks as fitness and leverages evolutionary algorithms to guide the per-layer KV cache compression. Specifically, in a language model with $L$ transformer layers, we denote the KV cache budget of layer $i$ as $k_i \in \mathbb{N}, \forall i \in \{1,\dots, L\}$. Given a set of candidate compression schemes $\mathbb{S}$ produced by an evolutionary algorithm for a downstream task $f(\cdot)$, we aim to identify the optimal scheme $S^*$ that maximizes task performance while minimizing deviation from the target average KV cache budget $c$:

\begin{align}
S^{*}&= \underset{S\in\mathbb{S}}{\arg\max}\;
      f(S)\,\bigl(1+\lambda\,\textproc{CacheScore}(S,c)\bigr) \nonumber \\ &\quad  {s.t.} \quad \frac{1}{L}\sum_{i=1}^L k_i \leq c
\end{align}

where $f(S)$ is the downstream-task performance obtained with compression scheme $S \in \mathbb{S}$, and the hyper-parameter $\lambda>0$ balances raw performance against cache efficiency.
Given the wide variety and differing value ranges of downstream performance metrics (e.g., accuracy, F1, ROUGE), we adopt a cache-efficiency term that is directly weighted against task performance to ensure comparability.
The cache-efficiency term $\textproc{CacheScore}(S,c)\in[0,1]$ assigns a lower value to schemes whose average per-layer cache budget $\bar{k}=\frac{1}{L}\sum_{i=1}^{L}k_i^{(S)}$ exceeds the target budget $c$, while applying a smooth discount to those that stay within the target:

\begin{equation}
\scalebox{0.85}{$
    \textproc{CacheScore}(S,c)=
    \begin{cases}
    \max\left(0, 1-\frac{\bar{k}-c}{c} \right), & \text{if } \bar{k} > c \\ 
1 - \gamma \left(1 - \frac{\bar{k}}{c} \right), & \text{if } \bar{k} \leq c  
\end{cases}
    $}
\end{equation}
where $\gamma\in(0,1]$ is a smoothing factor. 
Thus, the objective favors compression schemes that (i) deliver strong task performance and (ii) keep their average KV cache budgets close to, or below, the desired budget.

\paragraph{Grouping of KV Cache Budgets}
To improve optimization efficiency, we introduce a group size parameter $n_g$ to partition the KV cache budgets $\mathbf{K} = \{k_1, k_2, \dots, k_L\}$ into $J = \lceil L / n_g \rceil$ groups, denoted as $G = \{g_1, g_2, \dots, g_J\}$. Each group $g_j$ contains a contiguous subset of cache budgets defined as
$g_j = \{k_{(j-1) \cdot n_g + 1},\; k_{(j-1) \cdot n_g + 2},\; \dots,\; k_{\min(j \cdot n_g,\; L)}\}, \forall j \in \{1, 2, \dots, J\}.
$
For simplicity, we assume that the total number of layers $L$ is divisible by the group size $n_g$, such that $L = J \cdot n_g$. Under this formulation, candidate compression schemes $\mathbb{S}$ are applied at the group level and denoted by $\mathbb{S}_g$. The optimal scheme selected for each group, based on downstream task performance, is denoted by $S_g^*$. This group-wise formulation significantly reduces the search space and facilitates more stable optimization dynamics during the evolutionary search process.

\begin{algorithm}[!h]
  \caption{EvolKV — Evolutionary and Group-wise KV Cache Budget Optimization}
  \label{alg:evolkv}
  \scriptsize
  \begin{algorithmic}[1]
    \Require
      Target average KV cache budget $c$; cache budget efficiency weight $\lambda$;
      smoothing factor $\gamma$; group size $n_g$; the number of model layers $L$;
      max iterations $M$;
      \textproc{CacheScore} function;
      downstream task scorer $f(\cdot)$; evolutionary optimizer $\mathcal{A}$
    \Ensure
      Globally optimal group KV cache budgets $G^*$

    \State Initialize KV cache budgets $\mathbf{K}\gets(c,\dots,c)\in\mathbb{N}^{L}$
    \State Partition $\mathbf{K}$ into $J=\lceil L/n_g\rceil$ groups $G=\{g_1,\dots,g_J\}$
    \State $G^{*}\gets G $ \Comment{initialize group KV cache budgets}
    \State $F_{\text{best}}\gets-\infty$ \Comment{initialize global best fitness}

    \For{$j\gets1$ \textbf{to} $J$} \Comment{optimize one group at a time}
      \State $\mathcal{A}.\textproc{Initialize}(g_j)$ \Comment{initialize parameters of $\mathcal{A}$}
      \For{$m\gets1$ \textbf{to} $M$}
          \State Obtain candidate group compression schemes $\mathbb{S}_g$ from $\mathcal{A}$
          
          \State Evaluate the fitness $r$ of each $S_g\in \mathbb{S}_g$:
          \State $\quad \tilde{G} = G^* \text{ with } g_j := S_g$
          \State $\quad r \gets f(\tilde{G})(1+\lambda \textproc{CacheScore}(S_g, c))$
          \State \Comment{evaluate with $g_j$ of $G^*$ replaced by $S_g$, others fixed}
          \State Update $F_{\text{best}}$ with $r$, $G^*$ with $\tilde{G}$ if $r>F_\text{best}$
        \State Update the evolutionary optimizer $\mathcal{A}$ using $r$ and $S_g$
      \EndFor
    \EndFor
    \State \Return $G^{*}$
  \end{algorithmic}
\end{algorithm}


\paragraph{Iterations of Evolutionary Compression}
Our KV cache budget optimization is conducted in a group-wise manner, as shown in Algorithm~\ref{alg:evolkv}, proceeding sequentially from the bottom to the top layers. During the optimization of each group, the KV cache budgets of previously optimized groups are fixed to their respective optimal schemes $S_g^*$, while the remaining groups retain their initial values. If a candidate scheme $S_g$ achieves a higher fitness score $r$ than the current best, the KV cache budgets of the current group are updated accordingly. This process is repeated iteratively until all groups have been optimized.

\paragraph{KV Cache Budget Completion}\label{Budget-Completion}To ensure fairness in evaluation, we complete any KV cache budget optimization result whose total size deviates from the target. Specifically, we first compute the discrepancy between the achieved total KV cache budget $A = \sum_{i=1}^{L} k_i$ and the target total budget $T = c \cdot L$, denoted as $ {\Delta}_\text{cache}= T - A$. This discrepancy is then proportionally redistributed across layers based on their original share of $A$. The completed KV cache budgets $B = \{b_1, b_2, \dots, b_{L}\},\text{where } b_i = \left\lceil k_i + \frac{k_i}{A} \cdot  {\Delta}_\text{cache} \right\rceil, i \in \{1, 2, \dots, L\}$.

\section{Experiments}\label{sec:experiments}
\subsection{Experiment Settings}
\paragraph{Models}We employ two open-source models, Mistral-7B-Instruct\footnote{\url{https://huggingface.co/mistralai/Mistral-7B-Instruct-v0.2}}~\cite{jiang2023mistral7b} with 32K context length and Llama-3-8B-Instruct~\cite{grattafiori2024llama3herdmodels} with 8K context length. 
\paragraph{Datasets}Our proposed EvolKV is evaluated on four benchmarks: LongBench~\cite{bai2024longbenchbilingualmultitaskbenchmark}, GSM8K~\cite{cobbe2021trainingverifierssolvemath}, Needle-in-a-Haystack (NIAH)\footnote{\url{https://github.com/gkamradt/LLMTest_NeedleInAHaystack}}, and RULER~\cite{hsieh2024rulerwhatsrealcontext}. In RULER, we evaluate EvolKV on eleven sub-datasets across three major tasks: \textit{Retrieval} (Single NIAH with three sub-datasets, Multi-keys NIAH with three sub-datasets, Multi-queries NIAH, Multi-values NIAH), \textit{Aggregation} (CWE, FWE), and \textit{Multi-hop Tracing}. For LongBench, we select sixteen representative sub-datasets spanning six major task categories: \textit{single-document QA} (NarrativeQA~\cite{kočiský2017narrativeqareadingcomprehensionchallenge}, Qasper~\cite{dasigi2021datasetinformationseekingquestionsanswers}, MultiFieldQA-en), \textit{multi-document QA} (HotpotQA~\cite{yang2018hotpotqadatasetdiverseexplainable}, 2WikiMultihopQA~\cite{ho2020constructingmultihopqadataset}, MuSiQue~\cite{trivedi2022musiquemultihopquestionssinglehop}), \textit{summarization} (GovReport~\cite{huang2021efficientattentionslongdocument}, QMSum~\cite{zhong2021qmsumnewbenchmarkquerybased}, MultiNews~\cite{Multi-News}), \textit{few-shot learning} (TREC~\cite{Learning-Question-Classifiers}, TriviaQA~\cite{triviaqa}, SAMSum~\cite{SAMSum}), \textit{synthetic reasoning} (PassageCount, PassageRetrieval-en), and \textit{code completion} (LCC~\cite{guo2023longcoderlongrangepretrainedlanguage}, RepoBench-P~\cite{liu2023repobenchbenchmarkingrepositorylevelcode}). For detailed introduction of datasets, see Table~\ref{tab:task-info} in Appendix~\ref{app:downstream-task-info}.

\paragraph{Baselines and Settings}For a fair comparison with the current strong baselines, we keep the KV cache budget and all other hyperparameters identical in all evaluations.

\indent(1) StreamingLLM~\cite{xiao2024efficientstreaminglanguagemodels}. This method retains the original KV cache together with those from the most recent window and initial cache, preserving the KV cache at fixed positions in every layer.

(2) SnapKV~\cite{li2024snapkvllmknowslooking}. This method selects relevant KV cache from the preceding sequence based on the most recent query states (window query states), with a uniform cache budget applied across all layers. We employ SnapKV as the base method and apply the KV cache budgets optimized by EvolKV for downstream task inference and baseline comparison. Notably, when the optimized budgets are uniform across all layers, the resulting configuration reduces to standard SnapKV.

(3) PyramidKV~\cite{cai2024pyramidkvdynamickvcache}. A rule-based and pyramid-shaped strategy that progressively reduces the KV cache budget from the lower to higher layers.

\paragraph{Experimental Setup}
We apply Covariance Matrix Adaptation Evolution Strategy (CMA-ES~\cite{CMA-ES}) as our evolutionary optimizer. We set the window size to 32, kernel size to 7 and apply max pooling. For EvolKV, we fix the hyperparameters to $\lambda$ = 0.3, $\gamma$ = 0.2, learning rate of CMA-ES $\sigma$ = 0.3 and group size $n_g$ = 8. The population size of CMA-ES is calculated according to the following empirical formula: $4 + \lfloor3\cdot \ln(n_g)\rfloor$~\cite{ParameterSettingforMulticoreCMA-ES}. 

\subsection{Results}
\begin{table*}
  \centering 
  \setlength{\tabcolsep}{6pt} 
  \scalebox{0.54}{%
    \begin{tabular}{%
      c
      *{3}{c}  
      *{3}{c}  
      *{3}{c}  
      *{3}{c}  
      *{2}{c}  
      *{2}{c}  
      c       
    }
    \toprule
    \multirow{2}{*}{Method}
      & \multicolumn{3}{c}{\textbf{Single-Document QA}}
      & \multicolumn{3}{c}{\textbf{Multi-Document QA}}
      & \multicolumn{3}{c}{\textbf{Summarization}}
      & \multicolumn{3}{c}{\textbf{Few-shot Learning}}
      & \multicolumn{2}{c}{\textbf{Synthetic}}
      & \multicolumn{2}{c}{\textbf{Code}}
      & \multicolumn{1}{c}{\multirow{2}{*}{\textbf{Avg.}}} \\
    \cmidrule(lr){2-4}
    \cmidrule(lr){5-7}
    \cmidrule(lr){8-10}
    \cmidrule(lr){11-13}
    \cmidrule(lr){14-15}
    \cmidrule(lr){16-17}
      & \multicolumn{1}{c}{NrtvQA}    & \multicolumn{1}{c}{Qasper}   & \multicolumn{1}{c}{MF-en}
    & \multicolumn{1}{c}{HotpotQA} & \multicolumn{1}{c}{2WikiMQA} & \multicolumn{1}{c}{Musique}
    & \multicolumn{1}{c}{GovReport}& \multicolumn{1}{c}{QMSum}    & \multicolumn{1}{c}{MultiNews}
    & \multicolumn{1}{c}{TREC}     & \multicolumn{1}{c}{TriviaQA} & \multicolumn{1}{c}{SAMSum}
    & \multicolumn{1}{c}{PCount}   & \multicolumn{1}{c}{PRE}      & \multicolumn{1}{c}{Lcc}
    & \multicolumn{1}{c}{RB-P}     &  \\
    \midrule
     Full & 26.95 & 32.99 & 49.78 & 44.23 & 27.51 & 18.49 & 33.09 & 24.51 & 27.12 & 71.00 & 86.23 & 43.09 & 2.91 & 86.31 & 57.27 & 53.88 & 42.84 \\ \hline
     \multicolumn{18}{c}{\textit{KV Size = 128}} \\ \hline
     SnapKV & \underline{21.89} & 21.28 & 42.76 & 37.76 & 21.71 & 14.71 & 19.32 & 21.71 & \bfseries 21.28 & 49.00 & 83.62 & \underline{39.99} & \underline{2.66} & 68.05 & \underline{51.78} & \bfseries 48.36 & 35.37 \\ 
        PyramidKV & 20.40 & \underline{21.39} & \underline{43.95} & \underline{38.96} & \underline{23.80} & \underline{15.46} & \underline{19.69} & \bfseries 22.36 & \underline{21.23} & \underline{51.00} & \underline{84.77} & \bfseries 40.07 & 2.56 & \underline{72.20} & \bfseries 52.36 & \underline{47.62} & \underline{36.11} \\ 
        StreamingLLM & 17.28 & 13.21 & 27.11 & 30.82 & 21.94 & 11.87 & 15.48 & 19.37 & 17.98 & 44.00 & 80.22 & 37.32 & \bfseries 3.75 & 23.77 & 51.43 & 45.50 & 28.82 \\ 
       \rowcolor{gray!20} EvolKV & \bfseries 22.76 & \bfseries 22.59 & \bfseries 44.02 & \bfseries 39.47 & \bfseries 24.16 & \bfseries 15.64 & \bfseries 19.90 &  \underline{21.93} & 21.20 & \bfseries 52.00 & \bfseries 86.83 & 39.83 & 2.35 & \bfseries 74.81 & 51.64 & 47.05 & \bfseries 36.64 \\ \hline
        \multicolumn{18}{c}{\textit{KV Size = 256}} \\ \hline
        SnapKV &  \underline{22.81} & 24.18 & \underline{47.95} & 38.39 & \underline{23.22} & 15.31 & \underline{21.91} & \bfseries 23.13 & \underline{23.30} & 61.50 & \bfseries 85.90 & \bfseries 41.37 & 3.01 & \bfseries 84.81 & \bfseries 55.23 & \underline{51.16} & \underline{38.95}  \\ 
        PyramidKV  & 21.68 & \bfseries 24.90 & 47.59 & \underline{39.13} & 23.03 & \underline{16.74} & 21.62 & \underline{23.01} & 22.83 & \underline{62.50} & \underline{84.65} & \underline{40.71} & \underline{3.13} & \underline{82.65} & \underline{54.29} & 50.68 & 38.70  \\ 
        StreamingLLM& 19.43 & 15.31 & 27.97 & 31.57 & 21.78 & 11.30 & 18.04 & 19.18 & 19.94 & 51.00 & 80.75 & 39.60 & \bfseries 3.65 & 16.90 & 53.96 & 47.28 & 29.85 \\ 
        \rowcolor{gray!20} EvolKV & \bfseries 22.87  & \underline{24.50}  & \bfseries 48.28  & \bfseries 40.26  & \bfseries 25.74  & \bfseries 17.39  & \bfseries 22.33  & 22.39  & \bfseries 23.61  & \bfseries 65.50  & 84.57  & 40.66  & 2.85  & 79.35  & 54.10  & \bfseries 51.20  & \bfseries 39.10  \\ \hline
        \multicolumn{18}{c}{\textit{KV Size = 512}} \\ \hline
        SnapKV & \underline{24.68} & \underline{27.97} & \underline{48.80} & 40.32 & \underline{24.89} & \underline{16.99} & \underline{23.73} & \bfseries 23.57 & \underline{24.63} & \underline{67.00} & \underline{86.12} & 41.43 & 2.47 & \underline{88.06} & \underline{56.37} & \underline{52.80} & \underline{40.61}  \\ 
        PyramidKV & 24.39 & 27.49 & 48.78 & \underline{40.92} & 24.58 & 16.16 & 23.44 & \underline{23.48} & 24.05 & \underline{67.00} & 85.87 & 41.42 & 2.86 & 86.23 & 55.62 & 51.88 & 40.26 \\ 
        StreamingLLM & 21.39 & 16.36 & 30.75 & 30.89 & 22.20 & 10.95 & 21.53 & 20.02 & 23.10 & 61.50 & 81.86 & \bfseries 41.72 & \bfseries 3.14 & 18.57 & 55.16 & 48.65 & 31.74 \\ 
        \rowcolor{gray!20} EvolKV & \bfseries 24.89 & \bfseries 29.00 & \bfseries 49.78 & \bfseries 41.57 & \bfseries 26.27 & \bfseries 18.34 & \bfseries 24.41 & 23.18 & \bfseries 25.00 & \bfseries 68.00 & \bfseries 87.07 & \underline{41.64} & \underline{2.87} & \bfseries 89.74 & \bfseries 56.58 & \bfseries 52.85 & \bfseries 41.32 \\ \hline
       \multicolumn{18}{c}{\textit{KV Size = 1024}} \\ \hline
        SnapKV & \underline{25.46} & 29.15 & \bfseries 49.03 & \underline{41.58} & 25.30 & \bfseries 18.96 & \underline{26.19} & \bfseries 23.99 & \bfseries 25.99 & \underline{69.50} & \bfseries 86.63 & \bfseries 43.01 & 2.84 & \underline{89.21} & \bfseries 57.41 & 53.25 & \underline{41.72} \\ 
        PyramidKV & 25.45 & \underline{29.97} & 48.72 & 41.02 & \bfseries 25.85 & \underline{18.53} & 25.27 & 23.66 & 25.52 & 69.00 & \underline{86.31} & \underline{42.20} & 2.66 & 86.67 & 56.39 & \bfseries 53.38 & 41.29 \\ 
        StreamingLLM & 22.74 & 18.51 & 31.03 & 33.03 & 22.57 & 11.85 & 24.09 & 20.75 & 25.54 & 64.00 & 84.71 & 41.26 & \bfseries 3.49 & 22.40 & 55.89 & 50.99 & 33.30 \\ 
        \rowcolor{gray!20} EvolKV & \bfseries 25.63 & \bfseries 30.30 & \underline{48.96} & \bfseries 42.84 & \underline{25.78} & 18.21 & \bfseries 26.99 & \underline{23.79} & \underline{25.95} & \bfseries 70.00 & 86.09 & 41.74 & \underline{3.06} & \bfseries 89.82 & \underline{57.01} & \underline{53.26} & \bfseries 41.84  \\ \hline
        \multicolumn{18}{c}{\textit{KV Size = 2048}} \\ \hline
        SnapKV & \bfseries 26.29 & \bfseries 32.65 & \underline{49.09} & 41.70 & \underline{27.39} & \underline{18.49} & \underline{28.77} & \bfseries 24.42 & \underline{26.55} & 70.00 & \underline{86.27} & \bfseries 42.47 & \underline{2.79} & \underline{87.56} & \bfseries 57.42 & 53.41 & \underline{42.20} \\ 
        PyramidKV & 25.61 & 31.33 & 48.89 & \underline{41.90} & 26.64 & 17.65 & 28.09 & 23.86 & 26.52 & \bfseries 71.50 & \bfseries 86.30 & \underline{42.27} & 2.52 & 86.85 & \underline{57.41} & \bfseries 53.66 & 41.94 \\ 
        StreamingLLM & 22.28 & 23.08 & 35.22 & 33.66 & 22.90 & 13.47 & 26.85 & 20.95 & 26.45 & 66.00 & 85.68 & 41.95 & 2.40 & 25.75 & 57.13 & 52.17 & 34.75 \\ 
        \rowcolor{gray!20} EvolKV & \underline{26.13} & \underline{32.31} & \bfseries 49.18 & \bfseries 42.69 & \bfseries 27.63 & \bfseries 18.64 & \bfseries 29.11 & \underline{24.10} & \bfseries 26.72 & \underline{71.00} & 86.25 & 42.07 & \bfseries 2.85 & \bfseries 87.81 & 57.12 & \underline{53.63} & \bfseries 42.33 \\
    \bottomrule
    \end{tabular}%
  }
  \caption{Comparison of KV cache compression methods on Mistral-7B-Instruct across LongBench tasks. EvolKV outperforms all baseline methods on average across KV cache budgets ranging from 128 to 2048, and even surpasses the full model on certain tasks such as TriviaQA.}
  \label{tab:Mistral-7B-Instruct longbench}
\vspace{-1em}
\end{table*}

\subsubsection{Experiments on LongBench}
\paragraph{Settings}We evaluate EvolKV on LongBench, a comprehensive benchmark consisting of six major task categories designed to assess a model’s capacity for long-context understanding and reasoning. We first employ Mistral-7B-Instruct as the backbone model and use F1 score as the optimization objective. KV cache budgets are optimized using only 30 randomly sampled instances from NarrativeQA, under a target average cache budget of $c = 128$. The resulting allocation, illustrated in Figure~\ref{fig:sub1 mistral_ntrv-30_longbench_budgets}, is then extrapolated to target cache budgets of 256, 512, 1024, and 2048 using the method described in Section~\ref{Budget-Completion}, without further optimization. Following the same protocol, we conduct experiments on Llama-3-8B-Instruct, selecting six representative sub-datasets from LongBench—NarrativeQA, HotpotQA, QMSum, TREC, PassageRetrieval-en, and LCC—each with five randomly sampled instances. We perform EvolKV optimization with $c = 128$, and report the corresponding downstream task metrics applied in optimization in Table~\ref{tab:task-info} (Appendix~\ref{app:downstream-task-info}). The optimized KV cache budgets are subsequently extended to target average values of 256, 512, and 1024. For $c = 2048$, we conduct a separate optimization to obtain the dedicated cache budget allocation.

\paragraph{Results}Table~\ref{tab:Mistral-7B-Instruct longbench} reports the evaluation results on 16 LongBench sub-datasets using Mistral-7B-Instruct, with all training samples removed. Across all evaluated KV cache budgets, EvolKV consistently achieves the highest average performance, outperforming all rule-based baselines. Furthermore, on several sub-datasets, including MultiFieldQA-en, 2WikiMultihopQA, MuSiQue, TriviaQA, and PassageRetrieval-en, EvolKV not only remains competitive with the uncompressed full model but even surpasses it at certain KV cache budgets.
Table~\ref{tab:llama3-8b-instruct longbench} presents analogous results on Llama-3-8B-Instruct, again with training samples excluded. EvolKV demonstrates superior performance across all KV cache budgets. Remarkably, at a cache budget of 128, EvolKV outperforms the strongest baseline by 7.69 percentage points on the TREC subset, highlighting its strong adaptability to diverse downstream tasks.

\paragraph{Analysis}We conduct a detailed analysis of EvolKV's performance across the six major task categories in the LongBench, with the results presented in Figure~\ref{fig:mistral-llama_longbench_overall} and~\ref{fig:mistral-llama_longbench-six} (Appendix~\ref{app:Performance Results Across the Six Major Task Categories in LongBench}). EvolKV already surpasses all baselines under low-budget settings (e.g., $c=128$ and $ 256$) across multiple tasks: on Mistral-7B-Instruct it outperforms every baseline in single- and multi-document QA, and few-shot learning, while on Llama-3-8B-Instruct it achieves the top scores in few-shot learning and code completion. When the budget is relaxed to the range from 512 to 2048, EvolKV’s advantage becomes even more pronounced: on Mistral it exceeds the best baseline by up to 1.5 points in multi-document QA and surpasses the full model by 1.8 points in the synthetic task at $c=1024$, whereas on Llama it surpasses all baselines in few-shot learning, synthetic, and code tasks, achieving 1–3 point gains over the full model in code task. Notably, this advantage persists even under extreme compression, as EvolKV still outperforms the full model at $c=128$ (1.5\% of the context length), where all other baselines fall short. 

These results collectively demonstrate EvolKV’s strong and stable performance across both constrained and relaxed budget conditions, particularly in long-context scenarios. In contrast, PyramidKV’s pyramid-style allocation incurs significant losses on the code task, and StreamingLLM trails behind on nearly all tasks, reinforcing the limitation of static, rule-based allocation schemes. 
Notably, The KV cache budgets optimized by EvolKV at $c = 128$ generalize smoothly to larger average cache budgets ranging from 256 to 2048, suggesting that the method captures stable, task-aligned importance patterns rather than overfitting to a specific allocation regime. Even when the 30 optimization instances are excluded from evaluation, EvolKV continues to outperform all baselines, suggesting that its performance improvements stem from more effective KV cache budget allocation rather than memorization of training examples. These findings collectively highlight the superiority of EvolKV's adaptive and task-aware strategy over traditional heuristic approaches in real-world inference scenarios.

\begin{table*}
  \centering
  \setlength{\tabcolsep}{6pt} 
  \scalebox{0.54}{
    \begin{tabular}{%
      c
      *{3}{c}  
      *{3}{c}  
      *{3}{c}  
      *{3}{c}  
      *{2}{c}  
      *{2}{c}  
      c       
    }
    \toprule
    \multirow{2}{*}{Method}
      & \multicolumn{3}{c}{\textbf{Single-Document QA}}
      & \multicolumn{3}{c}{\textbf{Multi-Document QA}}
      & \multicolumn{3}{c}{\textbf{Summarization}}
      & \multicolumn{3}{c}{\textbf{Few-shot Learning}}
      & \multicolumn{2}{c}{\textbf{Synthetic}}
      & \multicolumn{2}{c}{\textbf{Code}}
      & \multicolumn{1}{c}{\multirow{2}{*}{\textbf{Avg.}}} \\
    \cmidrule(lr){2-4}
    \cmidrule(lr){5-7}
    \cmidrule(lr){8-10}
    \cmidrule(lr){11-13}
    \cmidrule(lr){14-15}
    \cmidrule(lr){16-17}
      & \multicolumn{1}{c}{NrtvQA}    & \multicolumn{1}{c}{Qasper}   & \multicolumn{1}{c}{MF-en}
    & \multicolumn{1}{c}{HotpotQA} & \multicolumn{1}{c}{2WikiMQA} & \multicolumn{1}{c}{Musique}
    & \multicolumn{1}{c}{GovReport}& \multicolumn{1}{c}{QMSum}    & \multicolumn{1}{c}{MultiNews}
    & \multicolumn{1}{c}{TREC}     & \multicolumn{1}{c}{TriviaQA} & \multicolumn{1}{c}{SAMSum}
    & \multicolumn{1}{c}{PCount}   & \multicolumn{1}{c}{PRE}      & \multicolumn{1}{c}{Lcc}
    & \multicolumn{1}{c}{RB-P}     &  \\
    \midrule
        Full & 25.51 & 31.49 & 39.80 & 43.62 & 35.96 & 21.39 & 28.74 & 23.19 & 26.79 & 73.85 & 90.50 & 42.89 & 4.18 & 68.21 & 58.89 & 53.59 & 41.79 \\ \hline
        \multicolumn{18}{c}{\textit{KV Size = 128}} \\ \hline
        SnapKV & \underline{21.96} & \bfseries 13.51 & 30.85 & 35.68 & \underline{29.14} & 19.21 & \bfseries 19.37 & \underline{21.63} & \underline{20.14} & 46.15 & \underline{88.32} & \underline{38.28} & 4.30 & \bfseries 68.21 & 57.33 & \underline{54.85} & 35.56 \\ 
        PyramidKV & \bfseries 22.25 & \underline{13.20} & \underline{31.54} & \underline{39.01} & 27.57 & \bfseries 20.18 & \underline{19.19} & \bfseries 21.91 & \bfseries 20.71 & \underline{50.26} & 87.34 & \bfseries 38.53 & 4.55 & \bfseries 68.21 &  \bfseries 57.55 & 54.25 & \underline{36.02} \\ 
        StreamingLLM & 18.91 & 7.79 & 20.90 & 33.79 & 24.85 & 14.94 & 16.37 & 20.41 & 18.49 & 46.15 & 74.40 & 35.78 & \underline{4.75} & 67.18 & 55.59 & 52.32 & 32.04 \\ 
        \rowcolor{gray!20} EvolKV & 21.18 & 13.12 & \bfseries 33.64 & \bfseries 40.70 & \bfseries 32.42 & \underline{19.81} & 17.29 & 21.26 & 19.03 & \bfseries 57.95 & \bfseries 89.30 & 37.38 & \bfseries 5.00 & \underline{67.86} & \underline{57.41} & \bfseries 55.11 & \bfseries 36.78 \\ \hline
        \multicolumn{18}{c}{\textit{KV Size = 256}} \\ \hline
        SnapKV & \bfseries 24.49 & \bfseries 18.37 & 33.36 & \bfseries 42.95 & \bfseries 34.05 & \underline{20.36} & \bfseries 20.62 & \bfseries 22.03 & \bfseries 22.51 & 59.49 & 89.61 & \bfseries 39.22 & 5.00 & \bfseries 68.21 & \underline{59.35} & \underline{56.27} & \underline{38.49} \\
        PyramidKV & \underline{24.14} & 15.75 & \bfseries 34.15 & 40.36 & 29.78 & \bfseries 21.22 & \underline{20.11} & 21.95 & \underline{22.30} & \underline{64.62} & \underline{90.03} & 39.16 & \bfseries 5.08 & \bfseries 68.21 & 59.21 & 53.79 & 38.12 \\
        StreamingLLM & 8.88 & 10.80 & 20.34 & 34.21 & 25.62 & 15.54 & 19.29 & 20.30 & 20.75 & 53.33 & 78.75 & \underline{39.20} & 4.83 & \underline{66.36} & 58.29 & 54.50 & 33.81 \\
        \rowcolor{gray!20} EvolKV & 22.68 & \underline{17.48} & \underline{34.09} & \underline{42.40} & \underline{32.92} & 19.91 & 19.34 & \underline{21.99} & 21.62 & \bfseries 67.18 & \bfseries 90.27 & 39.08 & \underline{5.06} & \bfseries 68.21 & \bfseries 59.71 & \bfseries 56.70 & \bfseries 38.67 \\ \hline
        \multicolumn{18}{c}{\textit{KV Size = 512}} \\ \hline
        SnapKV & \bfseries 25.26 & \underline{22.45} & \underline{36.84} & \underline{42.84} & \bfseries 36.13 & \underline{20.53} & \bfseries 22.38 & 22.44 & \bfseries 24.11 & \underline{69.74} & \bfseries 90.39 & \bfseries 40.40 & 4.98 & \bfseries 68.21 & \bfseries 60.20 & \underline{55.58} & \underline{40.16} \\
        PyramidKV & \underline{24.97} & 21.58 & 36.45 & \bfseries 43.36 & 32.97 & 19.55 & \underline{22.24} & \bfseries 22.64 & \underline{23.75} & 68.72 & 90.31 & \underline{40.30} & \underline{5.00} & \underline{67.95} & 58.72 & 54.61 & 39.57 \\
        StreamingLLM & 21.20 & 11.02 & 21.82 & 35.97 & 26.61 & 15.23 & 20.90 & 20.66 & 23.66 & 62.56 & 84.04 & 40.27 & 4.75 & 66.66 & 59.74 & 55.14 & 35.64 \\
        \rowcolor{gray!20} EvolKV & \bfseries 25.26 & \bfseries 22.91 & \bfseries 37.17 & 42.26 & \underline{35.99} & \bfseries 20.55 & 21.72 & \underline{22.45} & 23.32 & \bfseries 71.28 & \underline{90.36} & 39.45 & \bfseries 5.26 & \bfseries 68.21 & \underline{60.04} & \bfseries 58.12 & \bfseries 40.27 \\ \hline
        \multicolumn{18}{c}{\textit{KV Size = 1024}} \\ \hline
        SnapKV & \bfseries 25.16 & \underline{25.92} & \underline{38.39} & \underline{43.48} & \underline{34.98} & \bfseries 20.25 & \bfseries 24.05 & 22.09 & 25.21 & \underline{72.82} & \underline{90.43} & \bfseries 40.96 & 5.08 & \bfseries 68.21 & 59.79 & \underline{56.50} & \underline{40.83} \\
        PyramidKV & 24.90 & 24.82 & \bfseries 38.75 & 43.33 & 34.86 & \underline{20.16} & \underline{24.03} & \underline{22.86} & \underline{25.36} & \underline{72.82} & 90.34 & \underline{40.69} & \underline{5.12} & \bfseries 68.21 & 59.05 & 53.87 & 40.57 \\
        StreamingLLM & 21.43 & 15.37 & 26.18 & 36.26 & 28.41 & 15.56 & 23.26 & 21.24 & \bfseries 25.46 & 66.67 & 86.01 & 40.59 & 3.96 & \underline{68.03} & \bfseries 60.11 & 56.11 & 37.17 \\
        \rowcolor{gray!20} EvolKV & \underline{24.91} & \bfseries 27.31 & 37.41 & \bfseries 43.62 & \bfseries 36.03 & 20.09 & 23.44 & \bfseries 23.09 & 24.70 & \bfseries 74.36 & \bfseries 90.51 & 40.55 & \bfseries 5.28 & \bfseries 68.21 & \underline{59.90} & \bfseries 57.43 & \bfseries 41.05 \\ \hline
        \multicolumn{18}{c}{\textit{KV Size = 2048}} \\ \hline
        SnapKV & \bfseries 25.91 & \bfseries 29.45 & \underline{38.98} & \underline{43.64} & 35.55 & \bfseries 21.52 & \bfseries 25.83 & \underline{23.12} & 26.25 & \bfseries 73.85 & \bfseries 90.56 & \underline{41.82} & \bfseries 4.99 & \bfseries 68.21 & \underline{59.55} & \underline{55.03} &  \underline{41.52} \\ 
        PyramidKV & \underline{25.62} & 29.21 & 37.43 & 43.55 & \bfseries 36.77 & \underline{21.48} & 25.75 & 22.85 &  \underline{26.32} & \bfseries 73.85 & 90.34 & \bfseries 41.85 & 4.45 & \bfseries 68.21 & 59.47 & 54.15 & 41.33 \\
        StreamingLLM & 23.54 & 23.26 & 30.13 & 39.28 & 32.61 & 17.15 & 25.05 & 21.59 & 26.26 & \underline{70.77} & 89.78 & 41.50 & 4.62 & \underline{67.69} & \bfseries 60.01 & \bfseries 57.17 & 39.40 \\
        \rowcolor{gray!20} EvolKV & 25.52 & \underline{29.44} & \bfseries 39.60 & \bfseries 44.19 & \underline{36.58} & 20.91 & \underline{25.78} & \bfseries 23.34 & \bfseries 26.40 & \bfseries 73.85 & \underline{90.48} & 41.50 & \underline{4.91} & \bfseries 68.21 & 59.45 & 54.83 & \bfseries 41.56 \\
    \bottomrule
    \end{tabular}%
  }
  \caption{Comparison of KV cache compression methods on Llama-3-8B-Instruct across LongBench tasks. EvolKV outperforms all baselines on average across KV cache budgets from 128 to 2048, and even surpasses the full model on tasks like RepoBench-P under the 128 budget.}
  \label{tab:llama3-8b-instruct longbench}
\end{table*}

\subsubsection{Experiments on GSM8K}
\begin{figure}[ht]
  \centering
  \begin{subfigure}[t]{0.45\textwidth}
    \centering
    \includegraphics[width=\linewidth]{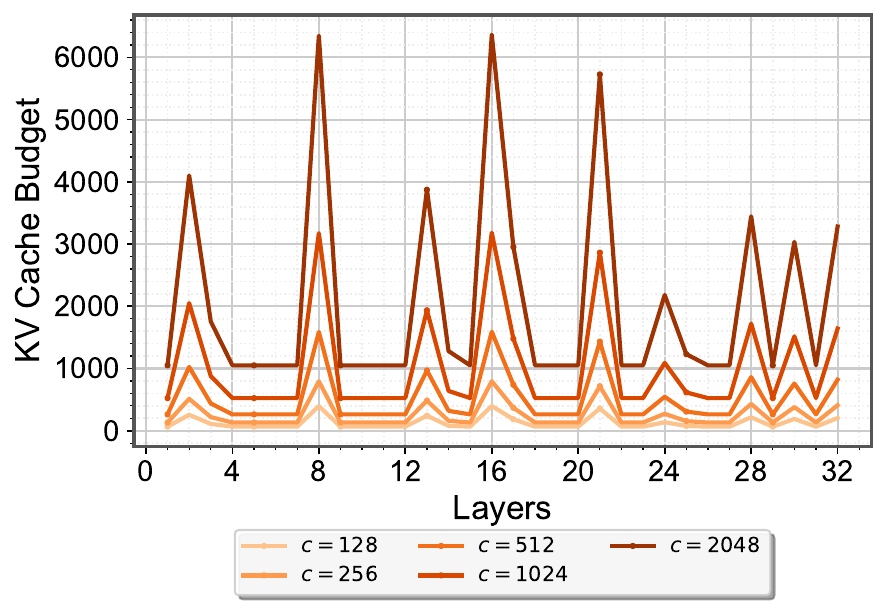}
    \caption{Mistral-7B-Instruct’s layer KV cache budgets optimized in 30 instances of NarrativeQA dataset.}
    \label{fig:sub1 mistral_ntrv-30_longbench_budgets}
  \end{subfigure}\hfill
  \begin{subfigure}[t]{0.45\textwidth}
    \centering
    \includegraphics[width=\linewidth]{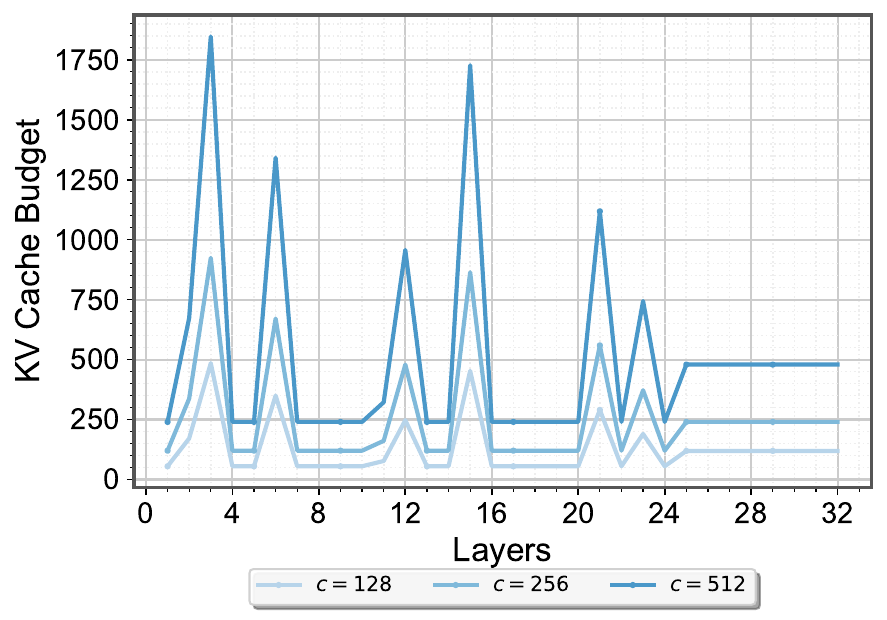}
    \caption{Llama-3-8B-Instruct's layer KV cache budgets optimized in 30 instances of GSM8K trainset data.}
    \label{fig:sub2 llama_gsm8k_layer_budget}
  \end{subfigure}
  \caption{Visualization of Mistral-7B-Instruct's layer KV cache budgets in LongBench and Llama-3-8B-Instruct' budgets in GSM8K.}
  \label{fig:Visualization-of-Longbench-GSM8K-layer-budgets}
\end{figure}

\paragraph{Settings}
We quantify the logical reasoning ability of EvolKV under different KV cache budgets on GSM8K. To eliminate prompt-format bias, all models receive identical few-shot chain-of-thought examples. For detailed prompt formatting and evaluation setup, we refer readers to the Qwen implementation\footnote{\url{https://github.com/QwenLM/Qwen/blob/main/eval/evaluate_chat_gsm8k.py}}. For EvolKV optimization, we randomly sample 30 instances from the GSM8K training set and perform KV cache budget optimization at $c = 128$, using accuracy as the objective metric. The resulting allocation is subsequently up-scaled to $c=256$ and $c=512$, without conducting additional searches. For Mistral-7B-Instruct, a separate optimization is performed under the setting $c=256$.
\paragraph{Results}Figure~\ref{fig:sub2 llama_gsm8k_layer_budget} presents the KV cache budget allocations optimized by EvolKV for Llama-3-8B-Instruct, and Table~\ref{tab:llama-mistral-gsm8k} reports the corresponding test set accuracies for both Llama-3-8B-Instruct and Mistral-7B-Instruct. Across all configurations, EvolKV consistently outperforms baseline methods on both models. Specifically, on Llama-3-8B-Instruct, it achieves substantial improvements over the strongest competitor, with accuracy gains of at least 7.28, 2.05, and 7.58 at KV cache budgets of 128, 256, and 512, respectively. Notably, EvolKV achieves 95.7\% of the full-model performance using a reduced cache budget ($c = 512$), significantly outperforming all baselines, whose best result reaches only 84.5\%.
\paragraph{Analysis}KV cache budgets optimized only at $c=128$ transfer effectively to larger budgets, indicating that the evolutionary search captures stable layer-importance signals rather than overfitting to a single setting. Notably, StreamingLLM performs poorly on this task, suggesting that fixed-position KV cache strategies are suboptimal for reasoning-oriented tasks. These results validate our central claim and extend our findings in LongBench: evolutionary, task-aware and layer-wise KV cache allocation unveils latent layer-specific cache requirements that fixed heuristics miss, yielding superior reasoning accuracy while retaining substantial memory savings.

\begin{table}[t]
  \centering 
  \setlength{\tabcolsep}{4pt}  
  \scalebox{0.84}{%
    \begin{tabular}{ccc}
    \toprule
    \multirow{2}{*}{Method}
    & \multicolumn{1}{c}{\textbf{Llama-3-8B-Instruct}} & \multicolumn{1}{c}{\textbf{Mistral-7B-Instruct}} \\
    & \multicolumn{1}{c}{Accuracy} & \multicolumn{1}{c}{Accuracy} \\
    \midrule
    Full & 67.85 & 50.80 \\
    \midrule
    \multicolumn{3}{c}{\textit{KV Size = 128}} \\
    \midrule
    SnapKV & 23.58 & 40.26 \\
    PyramidKV & \underline{40.71} & \underline{41.32} \\
    StreamingLLM & 3.64 & 2.58 \\
    \rowcolor{gray!20} EvolKV & \textbf{47.99} & \textbf{41.47} \\
    \midrule
    \multicolumn{3}{c}{\textit{KV Size = 256}} \\
    \midrule
    SnapKV & 44.50 & \underline{44.05} \\
    PyramidKV & \underline{49.05} & 43.59 \\
    StreamingLLM & 3.41 & 3.03 \\
    \rowcolor{gray!20} EvolKV & \textbf{51.10} & \textbf{44.81} \\
    \midrule
    \multicolumn{3}{c}{\textit{KV Size = 512}} \\
    \midrule
    SnapKV & 56.94 & \underline{46.02} \\
    PyramidKV & \underline{57.32} & \underline{46.02} \\
    StreamingLLM & 4.02 & 3.87 \\
    \rowcolor{gray!20} EvolKV & \textbf{64.90} & \textbf{46.70} \\
    \bottomrule
    \end{tabular}%
  }
  \caption{GSM8K results of Llama-3-8B-Instruct and Mistral-7B-Instruct under different KV cache budgets.}
  \label{tab:llama-mistral-gsm8k}
\end{table}

\begin{table*}
  \centering
  \setlength{\tabcolsep}{5pt} 
  \scalebox{0.63}{
    \begin{tabular}{%
      c
      c c c  
      c c c  
      c      
      c      
      c c    
      c      
      c      
    }
    \toprule
    \multirow{2}{*}{Method}
      & \multicolumn{8}{c}{\bfseries Retrieval}
      & \multicolumn{2}{c}{\bfseries Aggregation}
      & \multicolumn{1}{c}{\bfseries Multi-hop Tracing}
      & \multirow{2}{*}{Avg.} \\
    \cmidrule(lr){2-9} \cmidrule(lr){10-11} \cmidrule(lr){12-12}
      & S-NIAH-1
      & S-NIAH-2
      & S-NIAH-3
      & MK-NIAH-1
      & MK-NIAH-2
      & MK-NIAH-3
      & MQ-NIAH
      & MV-NIAH
      & CWE
      & FWE
      & VT
      & \\
    \midrule
        \multicolumn{13}{c}{\textbf{Mistral-7B-Instruct}} \\
        \midrule
        Full & 97.00  & 51.40  & 59.00  & 65.80  & 80.60  & 1.20  & 60.15  & 75.55  & 33.34  & 81.27  & 23.44  & 57.16 \\ \hline
        \multicolumn{13}{c}{\textit{KV Size = 128}} \\ \hline
        SnapKV & 29.40  & \underline{8.20}  & \underline{0.20}  & \underline{6.00}  & 3.80  & 0.00  & 0.05  & \underline{2.90}  & \underline{4.04}  & 30.53  & 7.16  & 8.39 \\ 
        PyramidKV & \underline{33.40}  & \bfseries 9.60  & 0.00  & \bfseries 7.40  & \underline{4.20}  & 0.00  & \underline{0.10}  & \bfseries 3.65  & \bfseries 4.48  & \underline{32.00}  & \underline{8.08}  & \underline{9.36} \\ 
        StreamingLLM & 0.20  & 0.80  & \bfseries 3.00  & 2.20  & 0.60  & 0.00  & \bfseries 1.50  & 1.65  & 0.18  & \bfseries 53.07  & 0.16  & 5.76 \\ 
        \rowcolor{gray!20} EvolKV & \bfseries 46.60  & 6.20  & 0.00  & \underline{6.00}  & \bfseries 8.60  & 0.00  & 0.00  & 1.45  & 3.36  & 31.80  & \bfseries 9.80  & \bfseries 10.35 \\ \hline
        \multicolumn{13}{c}{\textit{KV Size = 1024}} \\ \hline
        SnapKV & \underline{94.80}  & \underline{45.20}  & 2.80  & \underline{33.40}  & \underline{28.80}  & 0.00  & 9.05  & 9.20  & \underline{21.54}  & \underline{58.60}  & \underline{23.44}  & \underline{29.71} \\
        PyramidKV & \bfseries 95.60  & 44.60  & 2.20  & 30.60  & \bfseries 29.60  & 0.00  & \bfseries 10.15  & \bfseries 9.80  & 13.94  & 54.13  & \bfseries 23.60  & 28.57 \\
        StreamingLLM & 2.20  & 1.80  & \bfseries 5.60  & 4.00  & 2.80  & 0.00  & 2.40  & 2.45  & 0.16  & \bfseries 91.53  & 2.80  & 10.52 \\
        \rowcolor{gray!20} EvolKV & 94.60  & \bfseries 47.00  &\underline{3.00}  & \bfseries 40.20  & 24.80  & 0.00  & \underline{10.05}  & \underline{9.50}  & \bfseries 22.02  & 58.33  & 23.04  & \bfseries 30.23 \\
        \midrule
        \multicolumn{13}{c}{\textbf{Llama-3-8B-Instruct}} \\
        \midrule
        Full & 99.40  & 98.00  & 97.40  & 95.20  & 87.00  & 95.00  & 98.95  & 96.15  & 97.94  & 85.20  & 40.08  & 90.03 \\ \hline
        \multicolumn{13}{c}{\textit{KV Size = 128}} \\ \hline
        SnapKV & 75.80  & \underline{72.40}  & 1.40  & 29.80  & 0.40  & 0.00  & 20.65  & \underline{10.55}  & \bfseries 3.36  & \underline{48.00}  & \bfseries 10.24  & 24.78 \\ 
        PyramidKV & \underline{81.20}  & 69.60  & 1.40  & \bfseries 35.20  & \underline{0.60}  & 0.00  & \underline{23.15}  & \bfseries 10.75  & \underline{2.50}  & 46.93  & \underline{9.48}  & \underline{25.53} \\ 
        StreamingLLM & 1.60  & 1.40  & \bfseries 2.20  & 1.80  & \bfseries 1.00  & \bfseries 0.40  & 2.30  & 2.85  & 1.72  & \bfseries 92.07  & 2.68  & 10.00 \\ 
        \rowcolor{gray!20} EvolKV & \bfseries 90.00  & \bfseries 79.20  & \underline{1.80}  & \underline{33.40}  & \underline{0.60}  & 0.00  & \bfseries 26.90  & 10.35  & 1.88  & 44.40  & 8.52  & \bfseries 27.00 \\ \hline
        \multicolumn{13}{c}{\textit{KV Size = 1024}} \\ \hline
        SnapKV & \bfseries 100.00  & 93.00  & 2.20  & 82.20  & \underline{27.20}  & 6.00  & \underline{96.50}  & \underline{58.50}  & \bfseries 56.70  & \underline{77.87}  & \bfseries 51.88  & \underline{59.28} \\
        PyramidKV & \bfseries 100.00  & \bfseries 97.40  & 2.20  & \underline{82.60}  & 18.20  & 3.40  & 96.30  & 52.90  & 44.80  & 73.27  & 49.84  & 56.45 \\
        StreamingLLM & \underline{13.20}  & 12.80  & \bfseries 11.80  & 15.00  & 11.20  & \underline{11.00}  & 14.20  & 13.15  & 22.04  & \bfseries 90.93  & 17.68  & 21.18 \\
        \rowcolor{gray!20} EvolKV & \bfseries 100.00  & \underline{95.40}  & \underline{2.80}  & \bfseries 86.00  & \bfseries 34.60  & \bfseries 18.40  & \bfseries 97.70  & \bfseries 75.50  & \underline{53.94}  & 77.47  & \underline{49.92}  & \bfseries 62.88 \\
    \bottomrule
    \end{tabular}%
  }
  \caption{Comparison of KV cache compression methods on Mistral-7B-Instruct (top) and Llama-3-8B-Instruct (bottom) across RULER tasks. EvolKV consistently outperforms all baseline methods across KV cache budgets of 128 and 1024, achieving the highest average performance in both settings.}
  \label{tab:llama3-mistral-ruler}
\end{table*}

\begin{figure*}[t]
  \centering
   \begin{subfigure}[t]{0.3\textwidth}
    \centering
    \includegraphics[width=\linewidth]{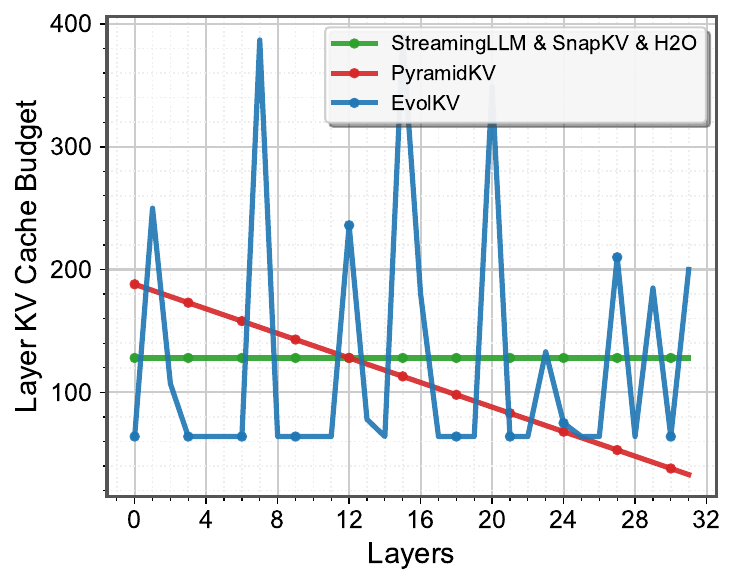}
    \caption{Comparison of EvolKV’s layer-wise KV cache budgets with existing KV cache compression methods.}
    \label{fig:sub Layer_KV_Budget_comparison}
  \end{subfigure}\hspace{0.02\textwidth}
  \begin{subfigure}[t]{0.3\textwidth}
    \centering
    \includegraphics[width=\linewidth]{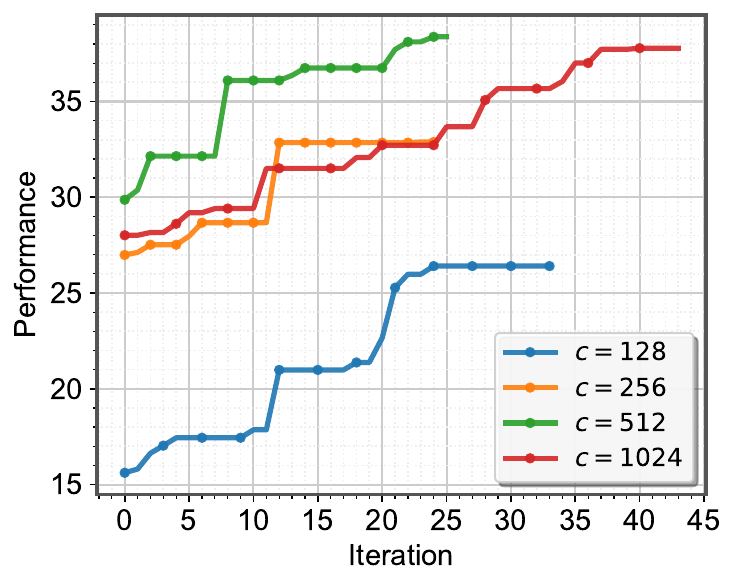}
    \caption{EvolKV optimizes layer-wise KV caches with target averages $c \in \{128, 256, 512, 1024\}$.}
    \label{fig:sub1 Visualization of Downstream Task Performance Optimization}
  \end{subfigure}\hspace{0.02\textwidth}
  \begin{subfigure}[t]{0.3\textwidth}
    \centering
    \includegraphics[width=\linewidth]{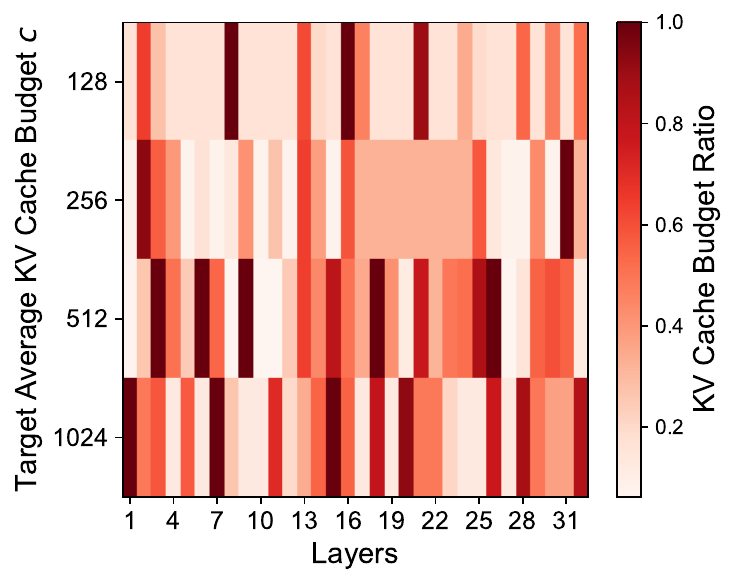}
    \caption{KV cache budget heatmap (compared to the maximum of layer KV cache budget).}
    \label{fig:sub2 Visualization of Downstream Task Performance Optimization}
  \end{subfigure}

  \caption{KV cache budget allocation of EvolKV—comparison, optimization trajectory, and allocation heatmap.}
  \label{fig:Visualization of Downstream Task Performance Optimization}
\end{figure*}

\subsubsection{Experiments on NIAH and RULER}
We evaluate the long-context retrieval capability of EvolKV alongside all baselines on NIAH. For optimization, EvolKV applies no more than 35 instances whose average scores on both Llama-3-8B-Instruct and Mistral-7B-Instruct are below 60, while the target KV cache budget $c$ is 128 and the optimization objective is the recall score. During the evaluation, the KV cache budget is fixed at 128. Figure~\ref{needle} (Appendix~\ref{app:Performance Results on NIAH}) presents the results: compared to the baselines, EvolKV achieves an improvement of over 4 percentage points on Llama-3-8B-Instruct and a substantial gain of more than 13 percentage points on Mistral-7B-Instruct. These results demonstrate that EvolKV effectively explores and leverages the latent layer-wise KV cache allocation of the model in long-context retrieval.

We further evaluate the KV cache allocations optimized in NIAH on the RULER benchmark, using Mistral-7B-Instruct and Llama-3-8B-Instruct with input context lengths set to 32K and 8K, respectively. Notably, the 1024 KV budget is extrapolated based on results from 128. As shown in Table~\ref{tab:llama3-mistral-ruler}, EvolKV consistently outperforms all baselines in average scores, with improvements of up to 0.99 points on Mistral-7B-Instruct and 3.6 points on Llama-3-8B-Instruct. These results further demonstrate the strong generalization, long-context retrieval and reasoning abilities of EvolKV, as the optimized KV budget can be effectively transferred to other benchmark evaluations, suggesting that EvolKV reveals the latent layer-wise allocation strategy.

\subsection{Analysis}
\paragraph{Downstream Task Performance during Optimization}
We randomly sample 30 instances from NarrativeQA as optimization data and conduct KV cache budget optimization experiments on Mistral-7B-Instruct, under target average cache budgets $c = 128$, $256$, $512$, and $1024$.
EvolKV can effectively optimize per-layer KV cache budget for downstream tasks during optimization. The results (depicted in Figure~\ref{fig:sub1 Visualization of Downstream Task Performance Optimization}) show a steady increase in model performance on the train data as the number of iterations grows. This suggests that the original, uniform per-layer KV cache budget allocation leaves substantial room for improvement on downstream tasks and that simple rule-based heuristics are insufficient to identify an optimal KV cache budget distribution.

\paragraph{Discussion on Optimized Budget Allocation} We optimize EvolKV on Mistral-7B-Instruct with randomly selected 30 NarrativeQA instances at $c=128,256,512$ and $1024$.
Our experiments reveal that EvolKV discovers totally distinct KV cache allocation patterns compared to heuristic approaches, as shown in Figure~\ref{fig:sub Layer_KV_Budget_comparison}. We observe consistent budget peaks at middle-model layers, suggesting these layers serve as computation bottlenecks for contextual inference. This non-intuitive pattern persists across tasks and budgets, showing that fixed heuristics inherently fail to capture transformer's dynamic layer utilization patterns. 
The allocation exhibits substantial variation across layers, depending on the target average cache budget $c$, as shown in Figure~\ref{fig:sub2 Visualization of Downstream Task Performance Optimization}. Under low-budget conditions (e.g., $c=12            8$), the optimization tends to concentrate cache budget resources on a small subset of layers. In contrast, higher budgets result in a more distributed allocation across the model. These patterns reveal that simple rule-based heuristics fail to capture the model’s latent cache-size preferences. Notably, as shown in Figure~\ref{fig:sub Layer_KV_Budget_comparison}, task-optimized KV cache allocations consistently diverge from rule-based heuristics, favoring non-uniform distributions that deviate from fixed or pyramidal rules. The cache-size requirements do not follow a pyramidal or monotonically decreasing pattern: several lower layers receive minimal allocation, while some higher layers are assigned significantly larger budgets, underscoring their greater relevance to downstream performance.

\begin{figure}[t]
  \centering
  \begin{subfigure}[t]{0.235\textwidth}
    \centering
    \includegraphics[width=\linewidth]{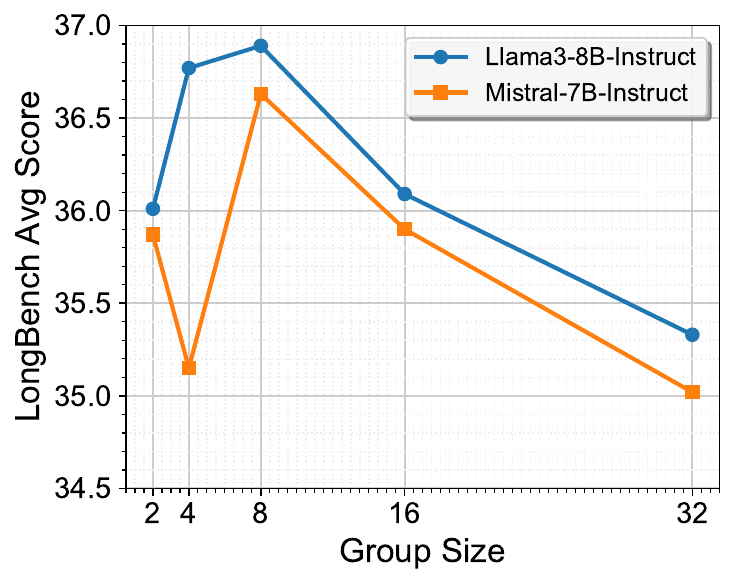}
    \caption{Comparison results of different group sizes. 
    }
    \label{fig:sub1 Layer_Frequency_Performance_Comparison}
  \end{subfigure}\hfill
  \begin{subfigure}[t]{0.235\textwidth}
    \centering
    \includegraphics[width=\linewidth]{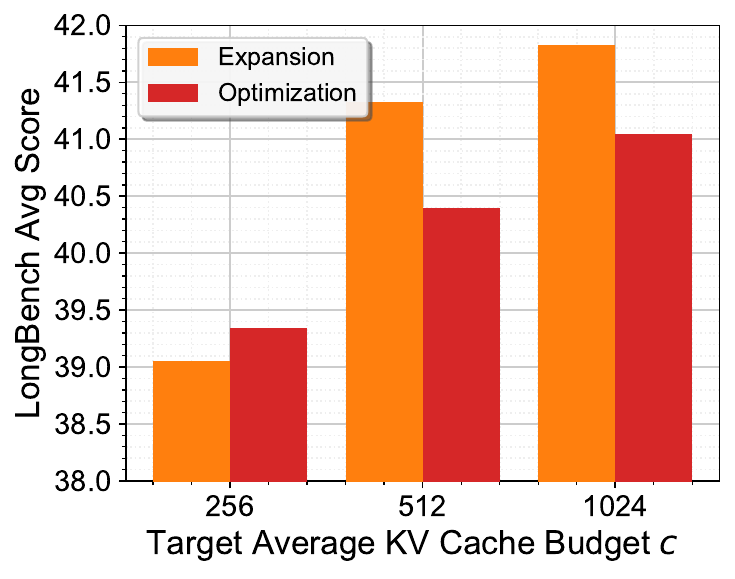}
    \caption{Comparison of expansion and optimization KV cache.}
    \label{fig:sub2 Optimization_vs_Expansion}
  \end{subfigure}
  \caption{Effects of layer grouping and KV cache expansion vs. KV cache optimization on LongBench.}
  \label{fig:Visualization-of-Group-size-KV-expansion}
\end{figure}

\paragraph{Effect of Group Size}
To construct the optimization dataset, we randomly sample five instances from each of the six LongBench subsets—NarrativeQA, HotpotQA, QMSum, TREC, PassageRetrieval-en, and LCC—for Llama-3-8B-Instruct. For Mistral-7B-Instruct, we sample 30 instances exclusively from NarrativeQA. We subsequently apply EvolKV to optimize the KV cache budgets for both models under a target average budget of $c=128$. During optimization, the group size $n_g$—i.e., the number of layers optimized concurrently—is set to 2, 4, 8, 16, and 32. The corresponding population sizes are determined using the empirical formula $4 + \lfloor 3 \cdot \ln(n_g) \rfloor$, yielding values of 6, 8, 10, 12, and 14, respectively. 
As shown in Figure~\ref{fig:sub1 Layer_Frequency_Performance_Comparison} and detailed results in Table~\ref{tab:ablation-mistral-llama-groupsize} (Appendix~\ref{app:Comparison Results of Different Group sizes}), downstream task performance generally improves as the group size increases, peaking at $n_g = 8$. Beyond this point, performance degrades notably. The inferior results with smaller group sizes likely stem from overfitting the layer-wise KV cache budget to limited optimization training data, while excessively large groups hinder effective budget allocation across layers. Accordingly, we select $n_g = 8$ as the optimal trade-off between performance and optimization efficiency.

\paragraph{Robustness Analysis of EvolKV}
Using Llama-3-8B-Instruct and a randomly sampled set of 30 NarrativeQA instances, we evaluate the stability of EvolKV at $c=128$. 
As detailed results reported in Table~\ref{tab:llama-3-8b-instruct-30-nrtv-3times-opti} (Appendix~\ref{app:Optimization Results Across Three Rounds}), after three optimization rounds, the grand average reaches 35.88. The deviation of each individual average from this overall average is within 0.1 and the standard deviation is 0.078, indicating that EvolKV exhibits consistent and stable behavior in all trials.
Additionally, We explore the impact of the selection of evolution training data on Mistral-7B-Instruct. Specifically, we randomly sample 30 instances drawn from six subsets (five items each) across the single- and multi-document QA categories. In parallel, we select another 30 instances, again five per subset, from six subsets (NarrativeQA, HotpotQA, QMSum, TREC, PassageRetrieval-en, and LCC).
The experimental results are summarized in Table~\ref{tab:ablation-mistral-train-data-select} (Appendix~\ref{Comparison Results of Different Selection of Training Data for Optimization}), where only the results obtained after removing the corresponding training data are reported. For the single- and multi-document QA group, we evaluate configurations with $c = 512$ and $1024$. While the overall performance is slightly lower than the main results, the $c = 512$ setting still outperforms the baselines, and the $c = 1024$ setting shows only a marginal drop. For the six-subsets group, we similarly examine $c = 1024$ and $2048$, both of which yield average scores that exceed the baselines. These results highlight the robustness of EvolKV and its strong adaptability across downstream tasks.

\paragraph{Generalization Analysis of EvolKV}
We optimize the layer-wise KV cache budgets of Mistral-7B-Instruct on a 30-sample subset of NarrativeQA with target average KV cache budget $c=256,512,\text{and }1024$. 
The results are shown in Figure~\ref{fig:sub2 Optimization_vs_Expansion} and the detailed results are shown in Table~\ref{tab:ablation-mistral-longbench-opt-or-expand} in Appendix~\ref{app:Comparison Results of KV Cache Budgets Expansion and Optimization}. As $c$ increases, direct cache expansion even outperforms direct optimization when $c=1024$ and $2048$. This finding suggests that when optimizing under strict budget constraints, EvolKV is able to effectively reveal the actual budget required for each layer and is able to try to generalize to higher target average budgets. Furthermore, we evaluate the KV cache budgets optimized at $c=128$ on the NIAH dataset by applying them to LongBench using Llama-3-8B-Instruct. The configuration for $c=256$ is derived from the $c=128$ allocation using the method described in Section~\ref{Budget-Completion}. As shown in Table~\ref{tab:ablation-niah-res-in-longbench-llama} (Appendix~\ref{app:Transfer Results of KV Cache Budget Optimization to Other Datasets}), EvolKV consistently outperforms baselines under both KV cache budget settings, indicating its ability to transfer cache budget allocations across datasets and demonstrating the ability of generalization.


\paragraph{Experiments on Models from Different Series}
We conduct KV cache budget optimization on the Qwen~\cite{bai2023qwentechnicalreport} series models at $c=128$ using 30 randomly sampled NarrativeQA instances. As shown in Table~\ref{tab:qwen-longbench-result}, with all training data excluded, EvolKV outperforms all other compression methods in terms of average score across both single- and multi-document QA tasks, demonstrating its consistent advantage across different model series. Notably, PyramidKV performs significantly worse than EvolKV, suggesting pyramidal budget allocation is suboptimal and highlighting EvolKV’s superior adaptability to downstream tasks.

\begin{table}
  \centering
  \setlength{\tabcolsep}{6pt} 
  \scalebox{0.55}{
    \begin{tabular}{%
      c
      *{3}{c}  
      *{3}{c}  
      c       
    }
    \toprule
    \multirow{2}{*}{Method}
      & \multicolumn{3}{c}{\textbf{Single-Document QA}}
      & \multicolumn{3}{c}{\textbf{Multi-Document QA}}
      & \multicolumn{1}{c}{\multirow{2}{*}{\textbf{Avg.}}} \\
    \cmidrule(lr){2-4}
    \cmidrule(lr){5-7}
      & \multicolumn{1}{c}{NrtvQA}    & \multicolumn{1}{c}{Qasper}   & \multicolumn{1}{c}{MF-en}
    & \multicolumn{1}{c}{HotpotQA} & \multicolumn{1}{c}{2WikiMQA} & \multicolumn{1}{c}{Musique}   &  \\
    \midrule
        \multicolumn{8}{c}{\textbf{Qwen2.5-1.5B-Instruct}} \\ \hline
        Full & 10.89 & 32.86 & 47.66 & 42.66 & 36.45 & 20.57 & 31.85 \\ \hline
        SnapKV & 9.13 & \underline{18.38} & \bfseries 37.91 & \underline{37.04} & 31.52 & \bfseries 14.87 & \underline{24.81} \\
        PyramidKV & \underline{9.23} & 17.40 & 35.85 & 36.11 & \underline{31.54} & 12.46 & 23.77 \\
        StreamingLLM & 6.99 & 17.23 & 23.97 & 28.35 & 29.82 & 8.20 & 19.09 \\
        \rowcolor{gray!20} EvolKV & \bfseries 9.34 & \bfseries 18.49 & \underline{37.80} & \bfseries 37.62 & \bfseries 32.98 & \underline{14.25} & \bfseries 25.08 \\ \hline
       \multicolumn{8}{c}{\textbf{Qwen2.5-3B-Instruct}} \\ \hline
       Full & 11.33 & 37.30 & 49.26 & 46.68 & 37.64 & 21.00 & 33.87 \\ \hline
        SnapKV & \bfseries 10.41 & \underline{23.58} & \underline{37.49} & \bfseries 39.85 & \underline{32.72} & \bfseries 14.65 & \underline{26.45} \\
        PyramidKV & 8.36 & 21.01 & 33.52 & 32.66 & 30.19 & 11.24 & 22.83 \\
        StreamingLLM & 8.11 & 19.98 & 23.52 & 32.60 & 28.95 & 10.26 & 20.57 \\
        \rowcolor{gray!20} EvolKV & \underline{9.68} & \bfseries 24.93 & \bfseries 38.70 & \underline{39.58} & \bfseries 33.43 & \underline{13.77} & \bfseries 26.68 \\
    \bottomrule
    \end{tabular}%
  }
  \caption{Comparison of KV cache compression methods on Qwen models across LongBench single- and multi-document tasks. EvolKV outperforms all baselines on average KV cache budget $c=128$.}
  \label{tab:qwen-longbench-result}
\end{table}

\section{Conclusion}
We introduce EvolKV, a task-driven framework that leverages evolution algorithm to optimize layer-wise KV cache budgets in LLMs. Unlike rule-based or heuristic methods, EvolKV directly maximizes downstream performance (e.g., accuracy, F1) without modifying model parameters and requires only a handful of labeled examples. Extensive experiments demonstrate that evolutionary, task-aware cache budget allocation uncovers latent layer-importance patterns overlooked by existing methods, and delivers state-of-the-art performance under stringent cache constraints and scaling gracefully to larger target cache budgets. Thus, EvolKV provides a practical, plug-and-play solution for efficient inference in downstream tasks. In the future, it would be valuable to investigate the robustness of KV cache compression under different tokenization schemes, as well as tokenization-free approaches~\cite{rust2022language, chai-etal-2024-tokenization, chai-etal-2024-autoregressive, cao2023unnatural}.

\section*{Limitations}
While EvolKV demonstrates strong performance on various downstream tasks, there is still room to explore budget allocation at the attention-head level. Future work will focus on combining downstream task performance with attention-head budgets.



\bibliography{EvolKV}
\bibliographystyle{acl_natbib}

\appendix

\section{Datasets} \label{app:downstream-task-info}
\begin{table*}[!ht]
  \centering
  \scalebox{0.76}{%
    \begin{tabular}{%
      l        
      l        
      c c c c c   
    }
    \toprule
      & \textbf{Datasets}
      & \textbf{Source}
      & \textbf{Avg len}
      & \textbf{Metric}
      & \textbf{Language}
      & \textbf{\#Sample} \\ \midrule

    \multirow{22}{*}{\rotatebox[origin=c]{90}{LongBench}}
        & \multicolumn{6}{l}{\textit{Single-Document QA}} \\ 
        & NarrativeQA        & Literature, Film   & 18,409 & F1           & English & 200 \\
        & Qasper             & Science            & 3,619  & F1           & English & 200 \\
        & MultiFieldQA-en    & Multi-field        & 4,559  & F1           & English & 150 \\ \cline{2-7}
        & \multicolumn{6}{l}{\textit{Multi-Document QA}} \\ 
        & HotpotQA           & Wikipedia          & 9,151  & F1           & English & 200 \\
        & 2WikiMultihopQA    & Wikipedia          & 4,887  & F1           & English & 200 \\
        & MuSiQue            & Wikipedia          & 11,214 & F1           & English & 200 \\ \cline{2-7}
        & \multicolumn{6}{l}{\textit{Summarization}} \\ 
        & GovReport          & Government report  & 8,734  & ROUGE-L      & English & 200 \\
        & QMSum              & Meeting            & 10,614 & ROUGE-L      & English & 200 \\
        & MultiNews          & News               & 2,113  & ROUGE-L      & English & 200 \\ \cline{2-7}
        & \multicolumn{6}{l}{\textit{Few-shot Learning}} \\ 
        & TREC               & Web question       & 5,177  & Accuracy (CLS)& English & 200 \\
        & TriviaQA           & Wikipedia, Web     & 8,209  & F1           & English & 200 \\
        & SAMSum             & Dialogue           & 6,258  & ROUGE-L      & English & 200 \\ \cline{2-7}
        & \multicolumn{6}{l}{\textit{Synthetic Reasoning}} \\ 
        & PassageCount       & Wikipedia          & 11,141 & Accuracy (EM)& English & 200 \\
        & PassageRetrieval-en& Wikipedia          & 9,289  & Accuracy (EM)& English & 200 \\ \cline{2-7}
        & \multicolumn{6}{l}{\textit{Code Completion}} \\ 
        & LCC                & GitHub             & 1,235  & Edit Sim     & Python/C\#/Java & 500 \\
        & RepoBench-P        & GitHub repository  & 4,206  & Edit Sim     & Python/Java & 500 \\ \midrule
    \multirow{14}{*}{\rotatebox[origin=c]{90}{RULER}}
        & \multicolumn{6}{l}{\textit{Retrieval}} \\ 
        & \multicolumn{6}{l}{\quad \textit{Single NIAH}} \\ 
        & S-NIAH-1 & – & – & Recall-based Accuracy & English & 500 \\
        & S-NIAH-2 & – & – & Recall-based Accuracy & English & 500 \\
        & S-NIAH-3 & – & – & Recall-based Accuracy & English & 500 \\ 
        & \multicolumn{6}{l}{\quad \textit{Multi-keys NIAH}} \\ 
        & MK-NIAH-1 & – & – & Recall-based Accuracy & English & 500 \\
        & MK-NIAH-2 & – & – & Recall-based Accuracy & English & 500 \\
        & MK-NIAH-3 & – & – & Recall-based Accuracy & English & 500 \\
        & Multi-queries NIAH (MQ-NIAH) & – & – & Recall-based Accuracy & English & 500 \\
        & Multi-values NIAH (MV-NIAH) & – & – & Recall-based Accuracy & English & 500 \\
        \cline{2-7}
        & \multicolumn{6}{l}{\textit{Aggregation}} \\ 
        & Common Words Extraction (CWE) & – & – & Recall-based Accuracy & English & 500 \\
        & Frequent Words Extraction (FWE) & – & – & Recall-based Accuracy & English & 500 \\
        \cline{2-7}
        & \multicolumn{6}{l}{\textit{Multi-hop Tracing}} \\ 
        & VT & – & – & Recall-based Accuracy & English & 500 \\
        \midrule
    \multicolumn{2}{l}{Needle-in-a-Haystack} & PaulGrahamEssays & –   & Recall   & English & –    \\ \midrule
    \multicolumn{2}{l}{GSM8K}                & Grade-school math word problems & 239 & Accuracy & English & 1319 \\ 
    \bottomrule
    \end{tabular}%
  }
  \caption{Datasets introduction. "Accuracy (CLS)" denotes classification accuracy; "Accuracy (EM)" denotes exact-match accuracy. "Recall-based Accuracy" denotes the proportion of reference strings that appear in the model output.}
  \label{tab:task-info}
\end{table*}
In this study, we apply LongBench~\cite{bai2024longbenchbilingualmultitaskbenchmark}, GSM8K~\cite{cobbe2021trainingverifierssolvemath}, Needle-in-a-Haystack\footnote{\url{https://github.com/gkamradt/LLMTest_NeedleInAHaystack}} and RULER~\cite{hsieh2024rulerwhatsrealcontext} to evaluate EvolKV. LongBench is a bilingual, multitask benchmark containing 21 datasets across six categories: single-document QA, multi-document QA, summarization, few-shot learning, synthetic tasks, and code completion. GSM8K (Grade School Math 8K) is a dataset comprising 8,792 high-quality grade school-level math word problems, including 7,473 for training and 1,319 for testing. The Needle-In-A-Haystack test is an evaluation method that measures a language model's ability to retrieve specific information ("needle") embedded within extensive, potentially distracting textual contexts ("haystack"). RULER is a long-context benchmark that is substantially more complex than NIAH, comprising four major tasks: retrieval, aggregation, multi-hop tracing, and question answering. Compared with NIAH, which mainly focuses on single retrieval accuracy, RULER emphasizes reasoning over long sequences and tests a model’s ability to handle compositional challenges such as integrating multiple retrieved pieces of evidence, following reasoning chains across documents, and synthesizing information over extended contexts. In this paper, we do not evaluate the question answering task, as one of its source datasets overlaps with LongBench. The detailed dataset information is presented in Table~\ref{tab:task-info}.

\section{Visualization results}

\subsection{Performance Results Across the Six Major Task Categories in LongBench}\label{app:Performance Results Across the Six Major Task Categories in LongBench}
\begin{figure*}[t]
  \centering
  \begin{subfigure}[t]{0.45\textwidth}
    \centering
    \includegraphics[width=\linewidth]{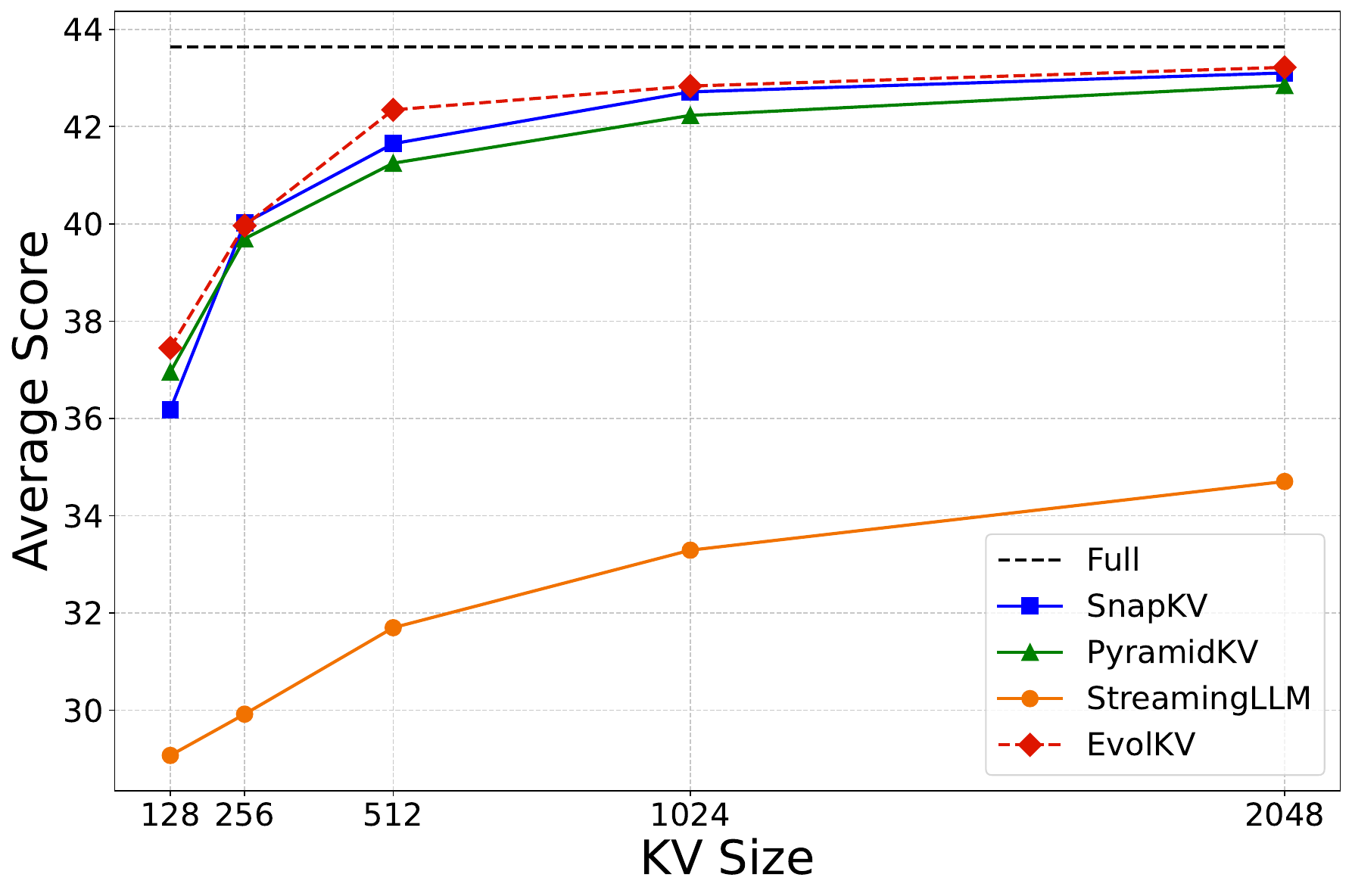}
    \caption{Average performance comparison across six major task categories in LongBench between baseline methods and our proposed EvolKV on Mistral-7B-Instruct.}
    \label{fig:sub1 mistral_longbench-overall}
  \end{subfigure}\hspace{0.04\textwidth}
  \begin{subfigure}[t]{0.45\textwidth}
    \centering
    \includegraphics[width=\linewidth]{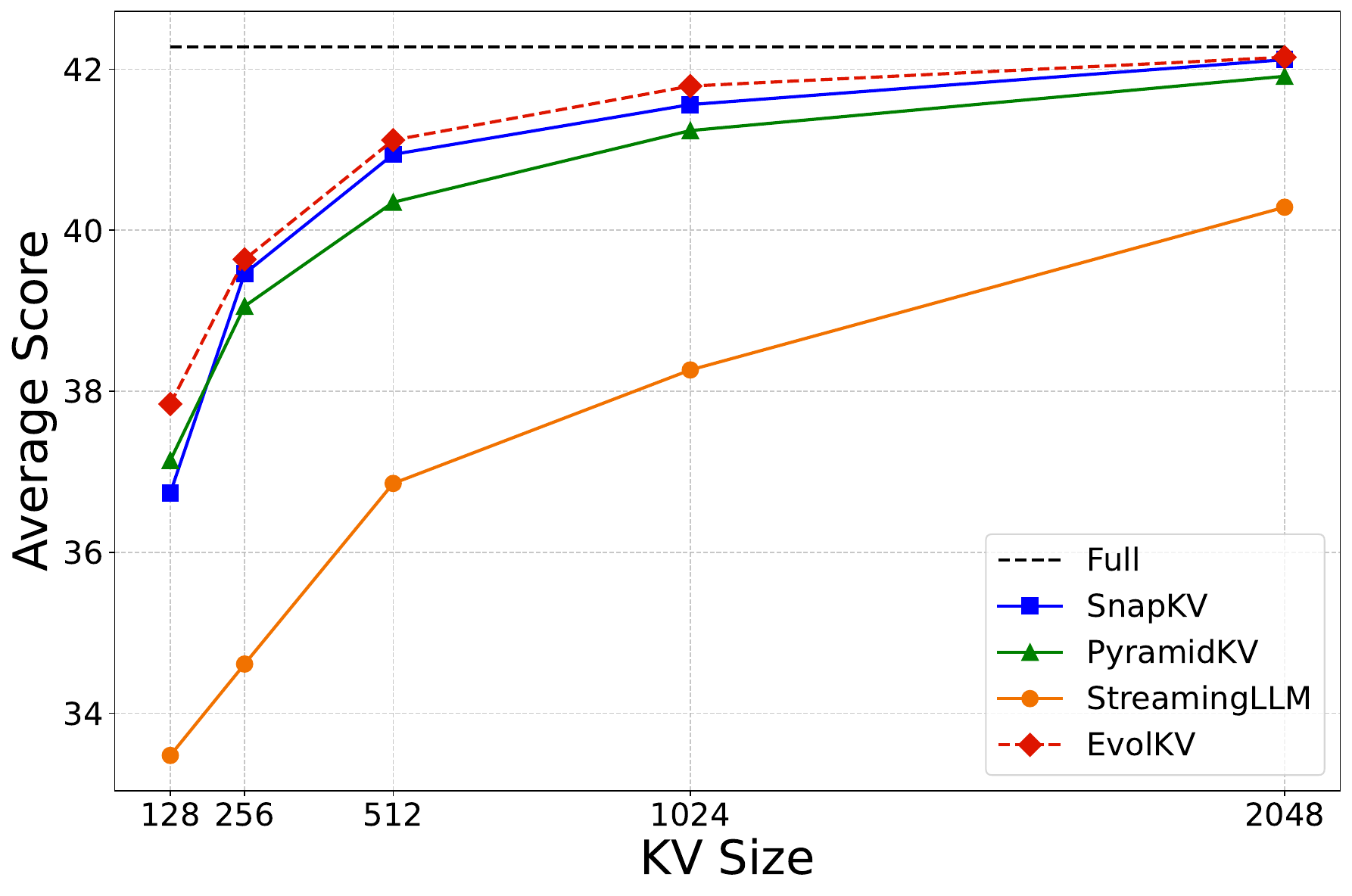}
    \caption{Average performance comparison across six major task categories in LongBench between baseline methods and our proposed EvolKV on Llama-3-8B-Instruct.}
    \label{fig:sub2 llama_longbench-overall}
  \end{subfigure}
  \caption{Average performance comparison across six major task categories in LongBench between baseline methods and our proposed EvolKV on Llama-3-8B-Instruct and Mistral-7B-Instruct.}
  \label{fig:mistral-llama_longbench_overall}
\end{figure*}

\begin{figure*}
  \centering
  \begin{subfigure}{1\textwidth}
    \centering
    \includegraphics[width=\linewidth]{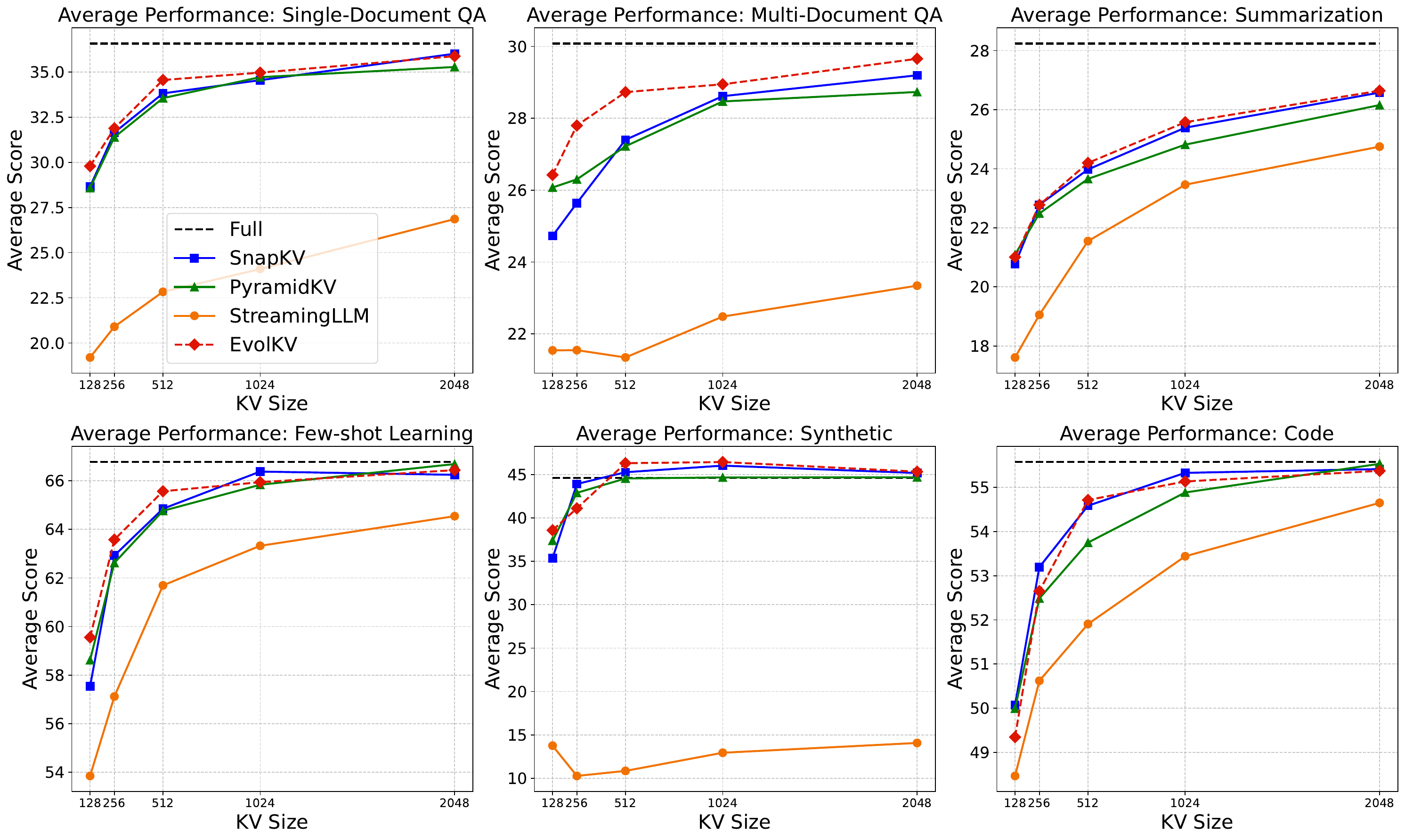}
    \caption{Comparison between baseline methods and our proposed EvolKV on Mistral-7B-Instruct across six major task categories in LongBench.}
    \label{fig:sub1 mistral_longbench-six}
  \end{subfigure}\hfill
  \begin{subfigure}{1\textwidth}
    \centering
    \includegraphics[width=\linewidth]{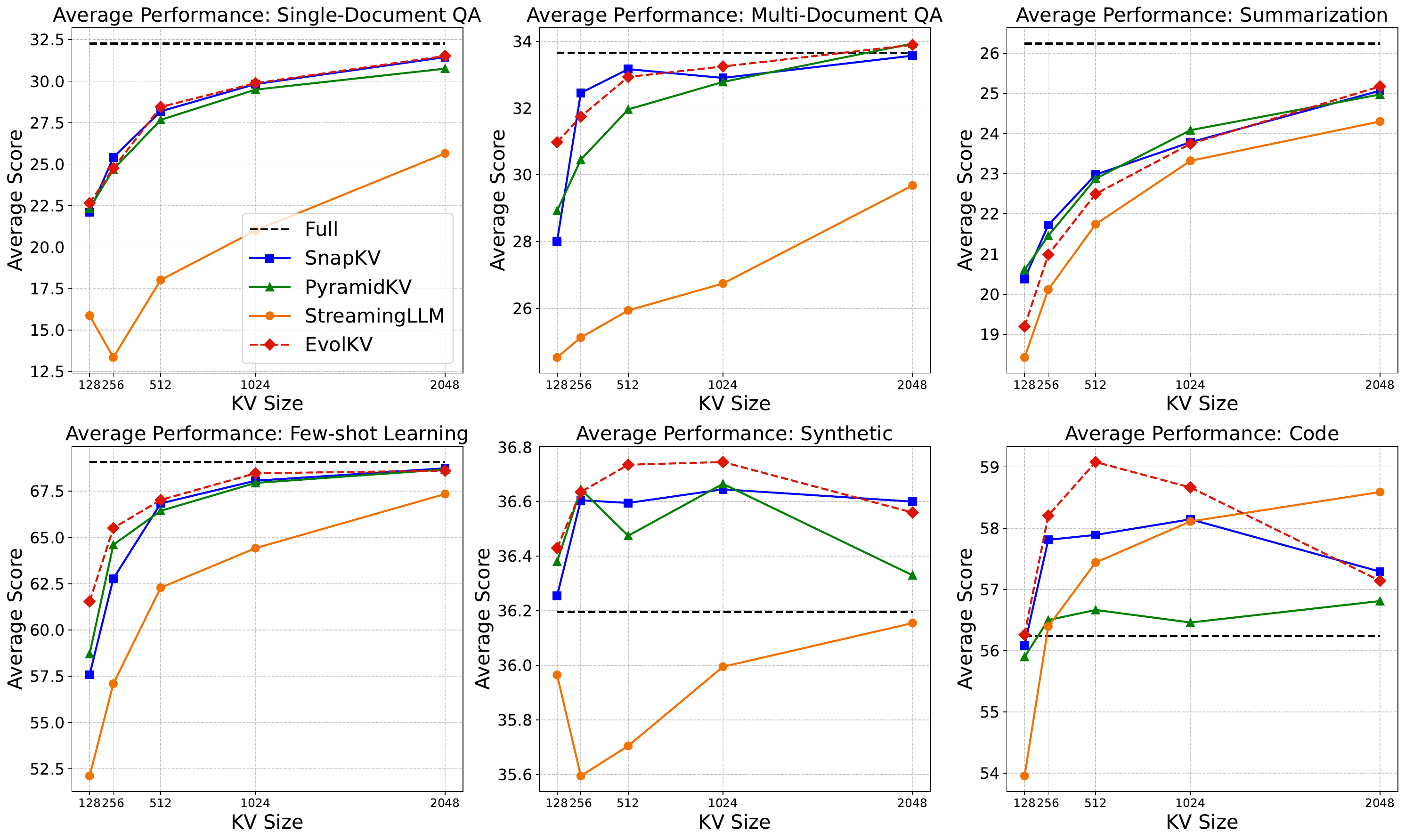}
    \caption{Comparison between baseline methods and our proposed EvolKV on Llama-3-8B-Instruct across six major task categories in LongBench.}
    \label{fig:sub2 llama_longbench-six}
  \end{subfigure}
  \caption{Performance comparison of baselines and EvolKV on Mistral-7B-Instruct and Llama-3-8B-Instruct across six LongBench task categories.}
  \label{fig:mistral-llama_longbench-six}
\end{figure*}

In this section, we present the performance of EvolKV and other baselines across the six major task categories in LongBench. As shown in Figure~\ref{fig:mistral-llama_longbench_overall}, EvolKV consistently outperforms all baselines in terms of average scores across the six categories. Figure~\ref{fig:mistral-llama_longbench-six} also demonstrates that EvolKV maintains an overall advantage across various task types. EvolKV consistently demonstrates superior performance across major task categories. Under both low and high KV cache budget settings, it maintains a leading position across various task types and even surpasses baseline methods under extremely constrained budgets in tasks such as synthetic, few-shot learning, and code. Notably, most of these results are based on the expanded KV cache budgets, indicating that EvolKV effectively discovers layer-wise cache allocation patterns tailored to downstream tasks, thereby highlighting its evolutionary and task-aware advantages.

\subsection{Performance Results on NIAH}\label{app:Performance Results on NIAH}
\begin{figure*}[htbp]
\centerline{\includegraphics[scale=0.48]{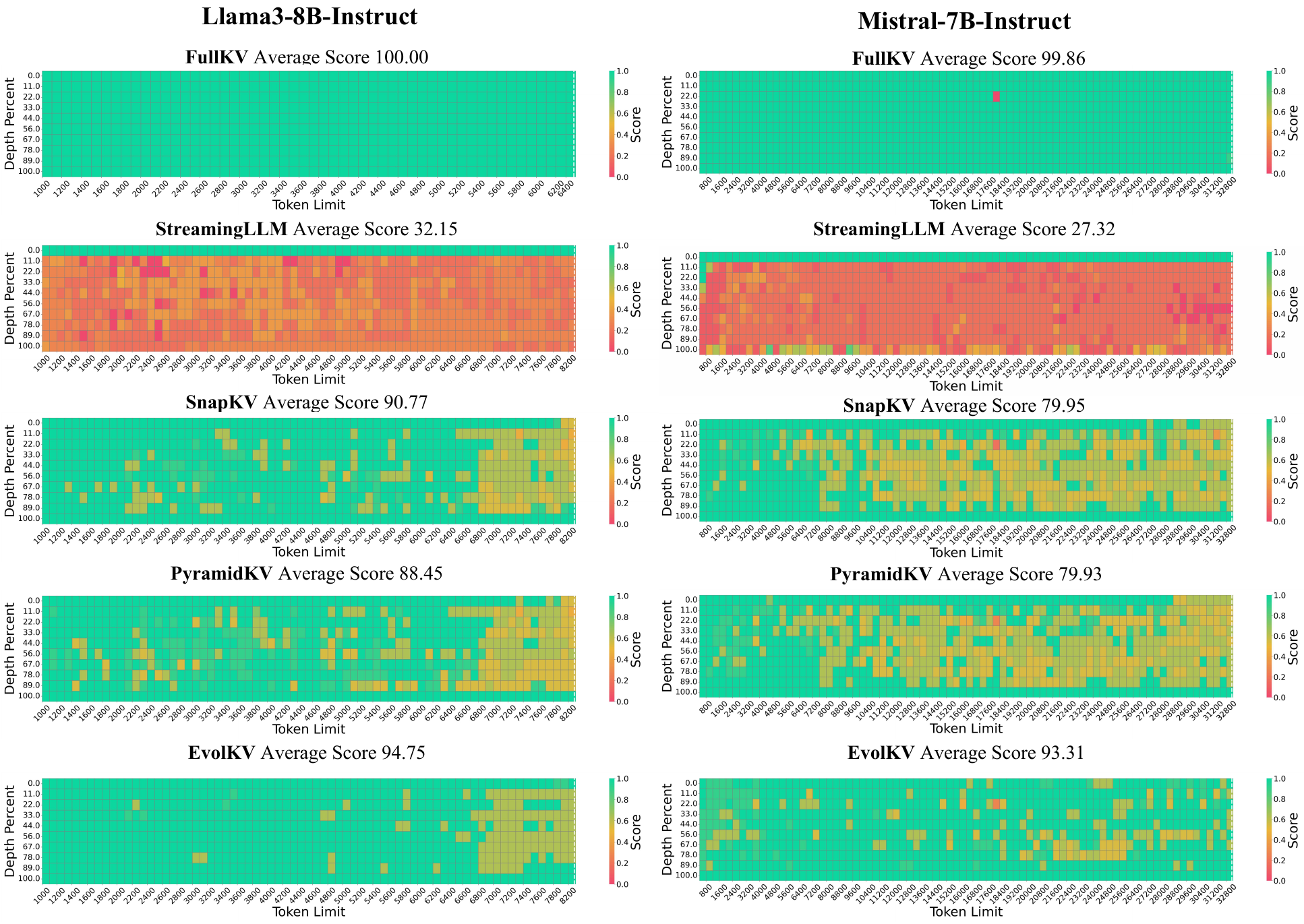}}
\caption{Evaluation of Llama-3-8B-Instruct and Mistral-7B-Instruct on NIAH at a KV cache budget of 128.}
\label{needle}
\end{figure*}

We evaluate EvolKV against other rule-based methods on NIAH at $c=128$, as shown in Figure~\ref{needle}. EvolKV consistently surpasses all baselines, achieving improvements of up to 13\% over the strongest competitor. In contrast, StreamingLLM exhibits weak performance on long-context retrieval, highlighting the inherent limitations of fixed-position methods and their tendency to lose critical contextual information. Notably, on Mistral-7B-Instruct with a 32K context length, SnapKV and PyramidKV already degrade significantly at 8K, whereas EvolKV maintains robust and superior retrieval ability across longer contexts.

\section{Detailed Results of Analysis}

\subsection{Comparison Results of Different Group sizes}
\label{app:Comparison Results of Different Group sizes}
Table~\ref{tab:ablation-mistral-llama-groupsize} presents the detailed downstream task performance across different group sizes. The overall performance peaks at $n_g=8$, which motivates our choice of setting the group size to 8 in practical experiments. Meanwhile, we argue that with only 30 training samples for optimization, overly small group sizes may cause EvolKV to overfit the limited data when allocating per-layer budgets, while overly large group sizes force EvolKV to handle too many layers simultaneously, making it difficult to perform fine-grained and effective allocation. Given that the models used in our experiments all have 32 layers, setting the group size to 8 serves as an optimal choice.

\begin{table*}
  \centering
  \setlength{\tabcolsep}{2pt} 
  \scalebox{0.64}{%
    \begin{tabular}{%
      c
      *{3}{S}  
      *{3}{S}  
      *{3}{S}  
      *{3}{S}  
      *{2}{S}  
      *{2}{S}  
      S       
    }
    \toprule
    \multirow{2}{*}{Method}
      & \multicolumn{3}{c}{\textbf{Single-Document QA}}
      & \multicolumn{3}{c}{\textbf{Multi-Document QA}}
      & \multicolumn{3}{c}{\textbf{Summarization}}
      & \multicolumn{3}{c}{\textbf{Few-shot Learning}}
      & \multicolumn{2}{c}{\textbf{Synthetic}}
      & \multicolumn{2}{c}{\textbf{Code}}
      & \multicolumn{1}{c}{\multirow{2}{*}{\textbf{Avg.}}} \\
    \cmidrule(lr){2-4}
    \cmidrule(lr){5-7}
    \cmidrule(lr){8-10}
    \cmidrule(lr){11-13}
    \cmidrule(lr){14-15}
    \cmidrule(lr){16-17}
      & \multicolumn{1}{c}{NrtvQA}    & \multicolumn{1}{c}{Qasper}   & \multicolumn{1}{c}{MF-en}
    & \multicolumn{1}{c}{HotpotQA} & \multicolumn{1}{c}{2WikiMQA} & \multicolumn{1}{c}{Musique}
    & \multicolumn{1}{c}{GovReport}& \multicolumn{1}{c}{QMSum}    & \multicolumn{1}{c}{MultiNews}
    & \multicolumn{1}{c}{TREC}     & \multicolumn{1}{c}{TriviaQA} & \multicolumn{1}{c}{SAMSum}
    & \multicolumn{1}{c}{PCount}   & \multicolumn{1}{c}{PRE}      & \multicolumn{1}{c}{Lcc}
    & \multicolumn{1}{c}{RB-P}     &  \\
    \midrule
        \multicolumn{18}{c}{\textbf{Llama-3-8B-Instruct}} \\ \hline
        EvolKV-$n_g2$ & 20.64 & \textbf{15.22} & \textbf{34.71} & 38.48 & 30.07 & 18.84 & \textbf{18.40} & 21.36 & 19.49 & 48.50 & 88.52 & 37.14 & 4.58 & \bfseries 69.00 & 56.99 & 54.18 & 36.01 \\
        EvolKV-$n_g4$ & \textbf{22.27} & 15.04 & 33.35 & 40.59 & 31.77 & 19.59 & 18.29 & \textbf{21.88} & \textbf{20.05} & 54.00 & \bfseries 89.33 & \bfseries 37.47 & 4.48 & 68.75 & 57.02 & 54.40 & 36.77  \\
        EvolKV-$n_g8$ & 21.4 & 13.12 & 33.64 & 41.3 & \bfseries 32.42 & \textbf{19.81} & 17.29 & 21.35 & 19.03 & \textbf{58.00} & 89.3 & 37.38 & \bfseries 5 & 68.67 & \bfseries 57.49 & \bfseries 55.11 & \bfseries 36.89  \\ 
        EvolKV-$n_g16$ & 21.62 & 13.31 & 34.26 & \textbf{41.64} & 31.76 & 18.37 & 17.07 & 21.22 & 19.2 & 48.5 & 88.3 & 36.81 & 4.65 & 68.75 & 57.13 & 54.87 & 36.09  \\ 
        EvolKV-$n_g32$ & 19.74 & 13.62 & 32.13 & 38.62 & 28.77 & 16.78 & 17.25 & 21.26 & 18.87 & 51.5 & 87.85 & 37.01 & 4.86 & \bfseries 69 & 55.32 & 52.71 & 35.33 \\ \hline
        \multicolumn{18}{c}{\textbf{Mistral-7B-Instruct}} \\ \hline
        EvolKV-$n_g2$ & 20.33 & 22.09 & \bfseries 45.64 & 38.19 & 23.01 & 15.82 & 19.65 & 21.66 & 21.1 & 49 & 85.17 & 40.34 & 2.71 & 69.14 & 52.35 & 47.67 & 35.87  \\
        EvolKV-$n_g4$ & 21.26 & 20.21 & 41.92 & 36.61 & 21.1 & 15.66 & 18.80 & 21.75 & 20.79 & 46 & 84.86 & \bfseries 40.57 & 2.19 & 72.24 & 50.88 & 47.6 & 35.15 \\
        EvolKV-$n_g8$ & 22.72 & \bfseries 22.59 & 44.02 & \bfseries 39.47 & \bfseries 24.16 & 15.64 & 19.9 & 21.93 & 21.2 & 52 & \bfseries 86.83 & 39.83 & 2.35 & \bfseries 74.81 & 51.64 & 47.05 & \bfseries 36.63 \\
        EvolKV-$n_g16$ & \bfseries 23.57 & 22.09 & 43.32 & 38.61 & 24.07 & \bfseries 16.31 & \bfseries 20.42 & \bfseries 22.09 & \bfseries 21.63 & \bfseries 58 & 84.7 & 39.16 & 2.89 & 55.5 & \bfseries 54.11 & \bfseries 48 & 35.90 \\
        EvolKV-$n_g32$ & 20.97 & 20.09 & 43.24 & 37 & 22.19 & 14.95 & 19.01 & 21.99 & 20.85 & 45.5 & 86.49 & 39.11 & \bfseries 3.77 & 67.37 & 50.64 & 47.08 & 35.02 \\
    \bottomrule
    \end{tabular}%
  }
  \caption{Comparison results of different group size ($n_g$). We use the empirical formula $4+\lfloor3\cdot \ln(n_g)\rfloor$ to obtain the population size.}
  \label{tab:ablation-mistral-llama-groupsize}
\end{table*}

\subsection{Optimization Results Across Three Rounds}
\label{app:Optimization Results Across Three Rounds}
We present in Table~\ref{tab:llama-3-8b-instruct-30-nrtv-3times-opti} the results of three optimization runs using EvolKV on the Llama-3-8B-Instruct model, conducted on 30 randomly sampled instances from the NarrativeQA dataset. The overall average score is 35.88, with a variance of 0.078, demonstrating the stable optimization performance of EvolKV.
\begin{table*}
  \centering
  \setlength{\tabcolsep}{2pt}
  \scalebox{0.7}{%
    \begin{tabular}{*{3}{S} *{3}{S} *{3}{S} *{3}{S} *{2}{S} *{2}{S} c}
    \toprule
      \multicolumn{3}{c}{\textbf{Single-Document QA}}
      & \multicolumn{3}{c}{\textbf{Multi-Document QA}}
      & \multicolumn{3}{c}{\textbf{Summarization}}
      & \multicolumn{3}{c}{\textbf{Few-shot Learning}}
      & \multicolumn{2}{c}{\textbf{Synthetic}}
      & \multicolumn{2}{c}{\textbf{Code}}
      & \multirow{2}{*}{\textbf{Avg.}}            
    \\
    \cmidrule(lr){1-3} \cmidrule(lr){4-6} \cmidrule(lr){7-9}
    \cmidrule(lr){10-12} \cmidrule(lr){13-14} \cmidrule(lr){15-16}
    \multicolumn{1}{c}{NrtvQA}    & \multicolumn{1}{c}{Qasper}   & \multicolumn{1}{c}{MF-en}
    & \multicolumn{1}{c}{HotpotQA} & \multicolumn{1}{c}{2WikiMQA} & \multicolumn{1}{c}{Musique}
    & \multicolumn{1}{c}{GovReport}& \multicolumn{1}{c}{QMSum}    & \multicolumn{1}{c}{MultiNews}
    & \multicolumn{1}{c}{TREC}     & \multicolumn{1}{c}{TriviaQA} & \multicolumn{1}{c}{SAMSum}
    & \multicolumn{1}{c}{PCount}   & \multicolumn{1}{c}{PRE}      & \multicolumn{1}{c}{Lcc}
    & \multicolumn{1}{c}{RB-P}     &  \\
    \midrule
        21.96 & 12.24 & \bfseries 35.47 & 37.92 & \bfseries 29.68 & \bfseries 20.87 & 18.76 & \bfseries 22.06 & \bfseries 20.16 & \bfseries 45 & \bfseries 88.72 & 38.4 & 3.98 & 68.27 & 56.96 & 55.19 & \bfseries 35.98 \\ 
        \bfseries 22.4 & 11.88 & 33.04 & \bfseries 39.96 & 29.27 & 19.97 & 18.93 & 21.83 & 19.78 & 43.5 & 87.53 & \bfseries 38.42 & \bfseries 4.5 & \bfseries 69 & \bfseries 58.39 & 55.39 & 35.86 \\ 
        21.34 & \bfseries 12.42 & 32.36 & 36.06 & 29.6 & 19.18 & \bfseries 19.01 & 21.37 & 19.99 & 44.5 & 88.2 & \bfseries 38.42 & 4.35 & 68.6 & 57.1 & \bfseries 60.11 & 35.79 \\
        
    \bottomrule
    \end{tabular}%
  }
  \caption{On Llama-3-8B-Instruct, 30 randomly selected instances from the NarrativeQA dataset are used to perform EvolKV layer-wise KV cache budget optimization at $c=128$ three times to evaluate stability.}
  \label{tab:llama-3-8b-instruct-30-nrtv-3times-opti}
\end{table*}

\subsection{Comparison Results of Different Selection of Training Data for Optimization}\label{Comparison Results of Different Selection of Training Data for Optimization}
We present in Table~\ref{tab:ablation-mistral-train-data-select} the results of EvolKV optimized with different training data configurations. The training data consists of two settings: the single-multi doc set, comprising 30 examples sampled from the single- and multi-document QA categories, and the six-subsets set, constructed from samples drawn from six datasets—NarrativeQA, HotpotQA, QMSum, TREC, PassageRetrieval-en, and LCC. The optimized results suggest that EvolKV holds a performance advantage, indicating its potential to maintain effective results across diverse downstream training data and demonstrating its strong adaptability.
\begin{table*}
  \centering
  \setlength{\tabcolsep}{2pt} 
  \scalebox{0.62}{%
    \begin{tabular}{%
      c
      *{3}{S}  
      *{3}{S}  
      *{3}{S}  
      *{3}{S}  
      *{2}{S}  
      *{2}{S}  
      S       
    }
    \toprule
    \multirow{2}{*}{Method}
      & \multicolumn{3}{c}{\textbf{Single-Document QA}}
      & \multicolumn{3}{c}{\textbf{Multi-Document QA}}
      & \multicolumn{3}{c}{\textbf{Summarization}}
      & \multicolumn{3}{c}{\textbf{Few-shot Learning}}
      & \multicolumn{2}{c}{\textbf{Synthetic}}
      & \multicolumn{2}{c}{\textbf{Code}}
      & \multicolumn{1}{c}{\multirow{2}{*}{\textbf{Avg.}}} \\
    \cmidrule(lr){2-4}
    \cmidrule(lr){5-7}
    \cmidrule(lr){8-10}
    \cmidrule(lr){11-13}
    \cmidrule(lr){14-15}
    \cmidrule(lr){16-17}
      & \multicolumn{1}{c}{NrtvQA}    & \multicolumn{1}{c}{Qasper}   & \multicolumn{1}{c}{MF-en}
    & \multicolumn{1}{c}{HotpotQA} & \multicolumn{1}{c}{2WikiMQA} & \multicolumn{1}{c}{Musique}
    & \multicolumn{1}{c}{GovReport}& \multicolumn{1}{c}{QMSum}    & \multicolumn{1}{c}{MultiNews}
    & \multicolumn{1}{c}{TREC}     & \multicolumn{1}{c}{TriviaQA} & \multicolumn{1}{c}{SAMSum}
    & \multicolumn{1}{c}{PCount}   & \multicolumn{1}{c}{PRE}      & \multicolumn{1}{c}{Lcc}
    & \multicolumn{1}{c}{RB-P}     &  \\
    \midrule
        \multicolumn{18}{c}{\textbf{Single-Multi Doc}} \\ \hline
        \multicolumn{18}{c}{\textit{KV Size = 512}} \\ \hline
        SnapKV & \bfseries 24.44 & \bfseries 28.14 & 48.09 & 40.16 & 25.09 & 16.45 & 23.73 & 23.57 & \bfseries 24.63 & 67 & 86.12 & 41.43 & 2.47 & 88.06 & 56.37 & \bfseries 52.8 & 40.53  \\
        PyramidKV & 23.2 & 27.83 & 48.15 & \bfseries 40.77 & 25.08 & 15.78 & 23.44 & 23.48 & 24.05 & 67 & 85.87 & 41.42 & 2.86 & \bfseries 86.23 & 55.62 & 51.88 & 40.17  \\
        StreamingLLM & 21.16 & 16.48 & 30.25 & 30.83 & 22.64 & 10.34 & 21.53 & 20.02 & 23.10 & 61.50 & 81.86 & 41.72 & \bfseries 3.14 & 18.57 & 55.16 & 48.65 & 31.68  \\
        \rowcolor{gray!20} EvolKV & 23.72 & 27.98 & \bfseries 48.41 & 40.69 & \bfseries 26.02 & \bfseries 17.13 & \bfseries 23.87 & \bfseries 23.64 & 24.62 & \bfseries 68.5 & \bfseries 86.52 & \bfseries 42.3 & 2.93 & 83.06 & \bfseries 56.99 & 52.44 & \bfseries 40.55 \\ \hline
        \multicolumn{18}{c}{\textit{KV Size = 1024}} \\ \hline
        SnapKV & \bfseries 25.23 & 29.45 & 48.54 & \bfseries 41.45 & 25.51 & 18.04 & \bfseries 26.19 & 23.99 & \bfseries 25.99 & 69.5 & \bfseries 86.63 & \bfseries 43.01 & 2.84 & \bfseries 89.21 & \bfseries 57.41 & 53.25 & \bfseries 41.64  \\
        PyramidKV & 24.27 & \bfseries 30.26 & 48.09 & 40.88 & \bfseries 26.39 & 17.88 & 25.27 & 23.66 & 25.52 & 69 & 86.31 & 42.2 & 2.66 & 86.67 & 56.39 & 53.38 & 41.18  \\
        StreamingLLM & 22.25 & 18.7 & 30.38 & 33.01 & 23.02 & 11.69 & 24.09 & 20.75 & 25.54 & 64 & 84.71 & 41.26 & \bfseries 3.49 & 22.4 & 55.89 & 50.99 & 33.26 \\
        \rowcolor{gray!20} EvolKV & 24.89 & 29.91 & \bfseries 48.84 & 41.28 & 26.03 & \bfseries 18.58 & 25.94 & \bfseries 24.2 & 25.92 & \bfseries 70 & 86.14 & 41.9 & 2.66 & 87.93 & 57.23 & \bfseries 53.64 & 41.57 \\ \hline
        \multicolumn{18}{c}{\textbf{Six-Subsets}} \\ \hline
        \multicolumn{18}{c}{\textit{KV Size = 1024}} \\ \hline
        SnapKV & \bfseries 25.49 & 29.15 & 49.03 & \bfseries 40.59 & 25.3 & 18.96 & \bfseries 26.19 & \bfseries 23.86 & 25.99 & 69.23 & \bfseries 86.63 & \bfseries 43.01 & 2.84 & \bfseries 88.94 & 57.24 & 53.25 & 41.61 \\
        PyramidKV & 24.22 & 29.97 & 48.72 & 40.03 & \bfseries 25.85 & 18.53 & 25.27 & 23.51 & 25.52 & 69.23 & 86.31 & 42.2 & 2.66 & 86.33 & 56.2 & \bfseries 53.38 & 41.12 \\
        StreamingLLM & 22.22 & 18.51 & 31.03 & 32.79 & 22.57 & 11.85 & 24.09 & 20.65 & 25.54 & 64.1 & 84.71 & 41.26 & \bfseries 3.49 & 21.44 & 55.71 & 50.99 & 33.18 \\
        \rowcolor{gray!20} EvolKV & 24.74 & \bfseries 30.53 & \bfseries 49.44 & 40.15 & 25.58 & \bfseries 19.6 & 26.12 & 23.61 & \bfseries 26.16 & \bfseries 70.26 & 86.53 & 42.06 & 2.52 & 88.27 & \bfseries 57.58 & 53.37 & \bfseries 41.66 \\ \hline
        \multicolumn{18}{c}{\textit{KV Size = 2048}} \\ \hline
        SnapKV & \bfseries 25.73 & 32.65 & \bfseries 49.09 & 40.78 & 27.39 & 18.49 & 28.77 & \bfseries 24.34 & 26.55 & 69.74 & 86.27 & \bfseries 42.47 & \bfseries 2.79 & \bfseries 87.24 & 57.24 & 53.41 & 42.06  \\
        PyramidKV & 22.14 & 23.08 & 35.22 & 32.94 & 22.9 & 13.47 & 26.85 & 20.85 & 26.45 & 66.15 & 85.68 & 41.95 & 2.4 & 24.87 & 56.95 & 52.17 & 34.63  \\
        StreamingLLM &25.32 & 31.33 & 48.89 & 40.92 & 26.64 & 17.65 & 28.09 & 23.7 & 26.52 & \bfseries 70.77 & \bfseries 86.3 & 42.27 & 2.52 & 86.51 & 57.24 & \bfseries 53.66 & 41.77 \\
        \rowcolor{gray!20} EvolKV & 25.28 & \bfseries 32.73 & 48.83 & \bfseries 40.94 & \bfseries 27.67 & \bfseries 18.97 & \bfseries 28.93 & 24.17 & \bfseries 26.94 & 69.74 & 86.27 & 42.12 & 2.64 & 86.94 & \bfseries 57.25 & 53.65 & \bfseries 42.07 \\
    \bottomrule
    \end{tabular}%
  }
  \caption{Comparison results of different selection of training data for EvolKV optimization. We report the results that removed training data.}
  \label{tab:ablation-mistral-train-data-select}
\end{table*}

\subsection{Comparison Results of KV Cache Budgets Expansion and Optimization}
\label{app:Comparison Results of KV Cache Budgets Expansion and Optimization}
Table~\ref{tab:ablation-mistral-longbench-opt-or-expand} compares the performance of two approaches: expanding KV cache budgets using the method described in Section~\ref{Budget-Completion} versus directly optimizing at the corresponding target average budget. It can be observed that when $c \ge 512$, the expansion-based method consistently outperforms direct optimization. This motivates our preference for using the expansion strategy to obtain target budgets rather than performing direct optimization. These results also demonstrate the simplicity of EvolKV’s optimization process—KV cache budget allocations optimized under low-budget settings can be directly extended to higher budgets without re-tuning, and even outperform directly optimized high-budget results in some tasks. This suggests that the low-budget phase enables more thorough exploration of task-aware per-layer KV cache allocation.

\begin{table*}
  \centering
  \setlength{\tabcolsep}{2pt} 
  \scalebox{0.65}{%
    \begin{tabular}{%
      c
      *{3}{S}  
      *{3}{S}  
      *{3}{S}  
      *{3}{S}  
      *{2}{S}  
      *{2}{S}  
      S       
    }
    \toprule
    \multirow{2}{*}{Method}
      & \multicolumn{3}{c}{\textbf{Single-Document QA}}
      & \multicolumn{3}{c}{\textbf{Multi-Document QA}}
      & \multicolumn{3}{c}{\textbf{Summarization}}
      & \multicolumn{3}{c}{\textbf{Few-shot Learning}}
      & \multicolumn{2}{c}{\textbf{Synthetic}}
      & \multicolumn{2}{c}{\textbf{Code}}
      & \multicolumn{1}{c}{\multirow{2}{*}{\textbf{Avg.}}} \\
    \cmidrule(lr){2-4}
    \cmidrule(lr){5-7}
    \cmidrule(lr){8-10}
    \cmidrule(lr){11-13}
    \cmidrule(lr){14-15}
    \cmidrule(lr){16-17}
      & \multicolumn{1}{c}{NrtvQA}    & \multicolumn{1}{c}{Qasper}   & \multicolumn{1}{c}{MF-en}
    & \multicolumn{1}{c}{HotpotQA} & \multicolumn{1}{c}{2WikiMQA} & \multicolumn{1}{c}{Musique}
    & \multicolumn{1}{c}{GovReport}& \multicolumn{1}{c}{QMSum}    & \multicolumn{1}{c}{MultiNews}
    & \multicolumn{1}{c}{TREC}     & \multicolumn{1}{c}{TriviaQA} & \multicolumn{1}{c}{SAMSum}
    & \multicolumn{1}{c}{PCount}   & \multicolumn{1}{c}{PRE}      & \multicolumn{1}{c}{Lcc}
    & \multicolumn{1}{c}{RB-P}     &  \\
    \midrule
        \multicolumn{18}{c}{\textit{KV Size = 256}} \\ \hline
        EvolKV.ex & 22.12  & \bfseries 24.50  &  \bfseries 48.28  &  \bfseries 40.26  & \bfseries 25.74  & \bfseries 17.39  &  \bfseries 22.33  & \bfseries 22.39  &  \bfseries 23.61  &  \bfseries 65.50  & 84.57  & 40.66  & 2.85  & 79.35  & 54.10  &  51.20  &  39.05 \\
        EvolKV.opt & \bfseries 24.53 & 24.14 & 47.14 & 40.04 & 24.54 & 17.23 & 21.62 & 22.85 & 23.09 & \bfseries 65.5 & \bfseries 86.5 & \bfseries 40.88 & \bfseries 3.69 & \bfseries 81.17 & \bfseries 54.85 & \bfseries 51.59 & \bfseries 39.34  \\ \hline
        \multicolumn{18}{c}{\textit{KV Size = 512}} \\ \hline
        EvolKV.ex &  24.99 &  \bfseries 29 &  \bfseries 49.78 &  \bfseries 41.57 &  \bfseries 26.27 & \bfseries 18.34 & \bfseries 24.41 & \bfseries 23.18 & \bfseries 25 & \bfseries 68 & \bfseries 87.07 & 41.64 & 2.87 & \bfseries  89.74 & \bfseries 56.58 & \bfseries 52.85 &  \bfseries 41.33 \\
        EvolKV.opt & \bfseries 25.69 & 27.59 & 48.79 & 41.16 & 25.97 & 17.16 & 22.65 & 22.9 & 24.21 & 67.5 & 86.27 & \bfseries 42.16 & \bfseries 3.55 & 82.77 & 56.04 & 51.99 & 40.40  \\ \hline
       \multicolumn{18}{c}{\textit{KV Size = 1024}} \\ \hline
        EvolKV.ex &  25.51 &  \bfseries 30.30 & 48.96 & \bfseries 42.84 & 25.78 & \bfseries 18.21 & \bfseries 26.99 & 23.79 &  \bfseries 25.95 &  \bfseries 70 & 86.09 & 41.74 & 3.06 & \bfseries 89.82 & 57.01 & 
 \bfseries 53.26 &  \bfseries 41.83  \\
        EvolKV.opt &  \bfseries 26.29 & 29.48 & \bfseries 49.15 & 40.36 & \bfseries 26.61 & 17.22 & 24.15 &  \bfseries 23.84 & 24.7 & 67 & \bfseries 86.45 &  \bfseries 42.69 & \bfseries 3.29 & 85.31 & \bfseries 57.06 & 53.16 & 41.05 \\
    \bottomrule
    \end{tabular}%
  }
  \caption{Comparison results of optimization and expansion. ".ex" denotes using the budget allocation optimized at $c=128$ to expand to target average KV cache budgets $c$; ".opt" denotes using EvolKV optimization at the corresponding $c$.}
  \label{tab:ablation-mistral-longbench-opt-or-expand}
\end{table*}

\subsection{Transfer Results of KV Cache Budget Optimization to Other Datasets}
\label{app:Transfer Results of KV Cache Budget Optimization to Other Datasets}
Table~\ref{tab:ablation-niah-res-in-longbench-llama} presents the evaluation results on the LongBench benchmark by directly applying the KV cache budget scheme optimized on the Needle-in-a-Haystack (NIAH) dataset using the Llama-3-8B-Instruct model. Notably, even without further adaptation, EvolKV consistently achieves the highest average performance across downstream tasks under low-budget constraints ($c=128$ and $256$), demonstrating its robust generalization capability in transferring optimized budget allocations from a single dataset to diverse task settings.
Furthermore, the budget configuration at $c=256$ is derived by directly expanding the allocation optimized under $c=128$, further reinforcing the scalability of EvolKV’s budget allocation results, as discussed in Section~\ref{app:Comparison Results of KV Cache Budgets Expansion and Optimization}. This not only emphasizes the method’s efficiency in low-resource settings but also highlights its potential for progressive extension to higher budgets without retraining, offering a compelling advantage in practical deployment scenarios.

\begin{table*}
  \centering
  \setlength{\tabcolsep}{2pt} 
  \scalebox{0.63}{%
    \begin{tabular}{%
      c
      *{3}{S}  
      *{3}{S}  
      *{3}{S}  
      *{3}{S}  
      *{2}{S}  
      *{2}{S}  
      S       
    }
    \toprule
    \multirow{2}{*}{Method}
      & \multicolumn{3}{c}{\textbf{Single-Document QA}}
      & \multicolumn{3}{c}{\textbf{Multi-Document QA}}
      & \multicolumn{3}{c}{\textbf{Summarization}}
      & \multicolumn{3}{c}{\textbf{Few-shot Learning}}
      & \multicolumn{2}{c}{\textbf{Synthetic}}
      & \multicolumn{2}{c}{\textbf{Code}}
      & \multicolumn{1}{c}{\multirow{2}{*}{\textbf{Avg.}}} \\
    \cmidrule(lr){2-4}
    \cmidrule(lr){5-7}
    \cmidrule(lr){8-10}
    \cmidrule(lr){11-13}
    \cmidrule(lr){14-15}
    \cmidrule(lr){16-17}
      & \multicolumn{1}{c}{NrtvQA}    & \multicolumn{1}{c}{Qasper}   & \multicolumn{1}{c}{MF-en}
    & \multicolumn{1}{c}{HotpotQA} & \multicolumn{1}{c}{2WikiMQA} & \multicolumn{1}{c}{Musique}
    & \multicolumn{1}{c}{GovReport}& \multicolumn{1}{c}{QMSum}    & \multicolumn{1}{c}{MultiNews}
    & \multicolumn{1}{c}{TREC}     & \multicolumn{1}{c}{TriviaQA} & \multicolumn{1}{c}{SAMSum}
    & \multicolumn{1}{c}{PCount}   & \multicolumn{1}{c}{PRE}      & \multicolumn{1}{c}{Lcc}
    & \multicolumn{1}{c}{RB-P}     &  \\
    \midrule
        \multicolumn{18}{c}{\textit{KV Size = 128}} \\ \hline
        SnapKV & 22.12 & \bfseries 13.51 & 30.85 & 35.98 & 29.14 & 19.21 & \bfseries 19.37 & 21.67 & 20.14 & 45.50 & 88.32 & 38.28 & 4.30 & \bfseries 69.00 & 57.39 & 54.85 & 35.60 \\
        PyramidKV & 22.42 & 13.20 & 31.54 & \bfseries 39.24 & 27.57 &  \bfseries 20.18 & 19.19 & \bfseries 21.92 & \bfseries 20.71 & \bfseries 50.00 & 87.34 & \bfseries 38.53 & 4.55 & \bfseries 69.00 & \bfseries 57.57 & 54.25 & 36.08 \\ 
        StreamingLLM & 18.96 & 7.79 & 20.90 & 33.13 & 24.85 & 14.94 & 16.37 & 20.45 & 18.49 & 45.50 & 74.40 & 35.78 & 4.75 & 68.00 & 55.65 & 52.32 & 32.02 \\ 
        \rowcolor{gray!20} EvolKV & \bfseries 22.78 & 13.02 & \bfseries 36.17 & 38.33 & \bfseries 31.08 & 19.89 & 18.22 & 21.57 & 19.92 & 48.00 & \bfseries 88.98 & 37.39 & \bfseries 5.08 & \bfseries 69.00 & 56.26 & \bfseries 55.65 & \bfseries 36.33 \\ \hline
        \multicolumn{18}{c}{\textit{KV Size = 256}} \\ \hline
        SnapKV & \bfseries 24.6 & \bfseries 18.37 & 33.36 & \bfseries 43.01 & \bfseries 34.05 & 20.36 & \bfseries 20.62 & 22.07 & \bfseries 22.51 & 60 & 89.61 & \bfseries 39.22 & 5 & \bfseries 69 & 59.47 & 56.27 & 38.60 \\ 
        PyramidKV & 24.2 & 15.75 & 34.15 & 40.48 & 29.78 & \bfseries 21.22 & 20.11 & 22 & 22.3 & \bfseries 65 & \bfseries 90.03 & 39.16 & \bfseries 5.08 & \bfseries 69 & 59.33 & 53.79 & 38.21 \\ 
        StreamingLLM & 18.9 & 10.8 & 20.34 & 33.54 & 25.62 & 15.54 & 19.29 & 20.31 & 20.75 & 53 & 78.75 & 39.2 & 4.83 & 67.2 & 58.39 & 54.5 & 33.81 \\ 
        \rowcolor{gray!20} EvolKV &  22.89 & 18.27 & \bfseries 35.6 & 41.76 & 33.48 & 20.54 & 20.47 & \bfseries 22.68 & 22.38 & 63.5 & 89.09 & 39.03 & 4.51 & \bfseries 69 & \bfseries 59.66 & \bfseries 56.52 & \bfseries 38.71 \\
    \bottomrule
    \end{tabular}%
  }
  \caption{Evaluation of Llama-3-8B-Instruct using KV cache budgets optimized on the Needle-in-a-Haystack from LongBench.}
  \label{tab:ablation-niah-res-in-longbench-llama}
\end{table*}

\begin{table}
  \centering
  \setlength{\tabcolsep}{4.2pt} 
  \scalebox{0.55}{
    \begin{tabular}{%
      c
      *{3}{c}  
      c
      *{3}{c}  
      c       
    }
    \toprule
    \multirow{2}{*}{Method}
      & \multicolumn{3}{c}{\textbf{Single-Document QA}}
      & \multicolumn{1}{c}{\multirow{2}{*}{\textbf{Avg.}}}
      & \multicolumn{3}{c}{\textbf{Multi-Document QA}}
      & \multicolumn{1}{c}{\multirow{2}{*}{\textbf{Avg.}}} \\
    \cmidrule(lr){2-4}
    \cmidrule(lr){6-8}
      & \multicolumn{1}{c}{NrtvQA}    & \multicolumn{1}{c}{Qasper}   & \multicolumn{1}{c}{MF-en} & 
    & \multicolumn{1}{c}{HotpotQA} & \multicolumn{1}{c}{2WikiMQA} & \multicolumn{1}{c}{Musique}   &  \\
    \midrule
        \multicolumn{9}{c}{\textbf{Mistral-7B-Instruct}} \\ \hline
        SnapKV & 20.13 & 21.41 & 41.67 & 27.74 & 36.48 & 21.53 & 14.43 & 24.15 \\ 
        PyramidKV &18.91 & \bfseries 21.67 & \bfseries 43.17 & 27.92 & \bfseries 37.65 & 23.47 & 15.12 & 25.41 \\ 
        StreamingLLM & 16.93 & 13.47 & 26.70 & 19.03 & 29.91 & 21.51 & 11.44 & 20.95 \\ 
        \rowcolor{gray!20} EvolKV & \bfseries 21.56 & 21.49 & 42.61 & \bfseries 28.55 & 36.79 & \bfseries 24.22 & \bfseries 16.37 & \bfseries 25.79 \\ \hline
       \multicolumn{9}{c}{\textbf{Llama-3-8B-Instruct}} \\ \hline
        SnapKV & \bfseries 21.48 & \bfseries 13.80 & 31.25 & 22.18 & 34.71 & 27.96 & 19.10 & 27.26 \\
        PyramidKV & 21.40 & 13.55 & 32.48 & 22.48 & \bfseries 38.05 & 26.75 & 19.63 & 28.14  \\ 
        StreamingLLM & 18.06 & 7.81 & 21.45 & 15.77 & 32.13 & 23.84 & 14.62 & 23.53  \\ 
        \rowcolor{gray!20} EvolKV & 21.24 & 13.13 & \bfseries 34.72 & \bfseries 23.03 & 37.86 & \bfseries 28.60 & \bfseries 20.33 & \bfseries 28.93 \\
    \bottomrule
    \end{tabular}%
  }
  \caption{Specialized task optimization results of EvolKV across LongBench single- and multi-document tasks. EvolKV outperforms all baselines on average KV cache budget $c=128$.}
  \label{tab:specialized-opt-result}
\end{table}

\subsection{Experiments on Specialized Task Optimization}
To evaluate the specialized task optimization performance of EvolKV, we perform KV cache budget optimization on both single-document and multi-document QA tasks, using 30 randomly sampled instances for each setting, with 10 instances per sub-task. As shown in Table~\ref{tab:specialized-opt-result}, after removing all training data, EvolKV achieves the highest average scores at $c=128$ on both tasks, surpassing all baselines and demonstrating its strong adaptability to downstream tasks.

\subsection{Evaluation of Inference Time and Memory Cost}
\begin{figure*}[t]
  \centering
  \begin{subfigure}[t]{0.47\textwidth}
    \centering
    \includegraphics[width=\linewidth]{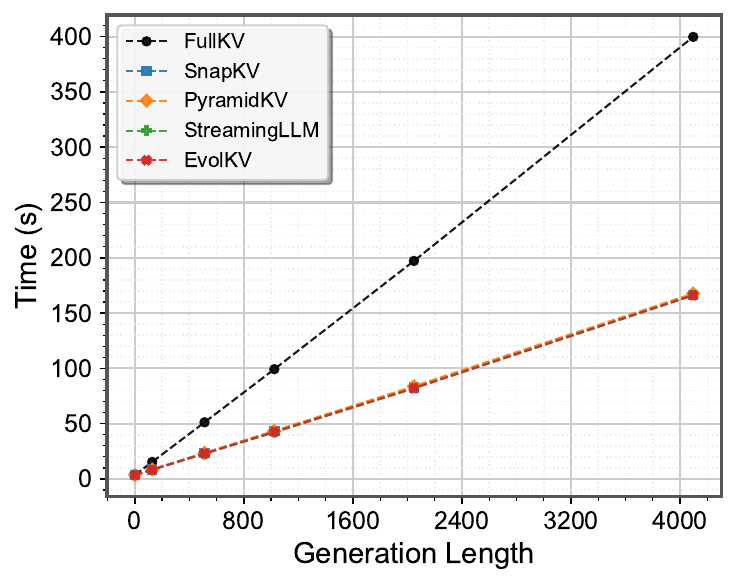}
    \caption{Comparison results of inference time.}
    \label{fig:sub1 decoding_res_compare}
  \end{subfigure}
  \hspace{0.005\textwidth}
  \begin{subfigure}[t]{0.47\textwidth}
    \centering
    \includegraphics[width=\linewidth]{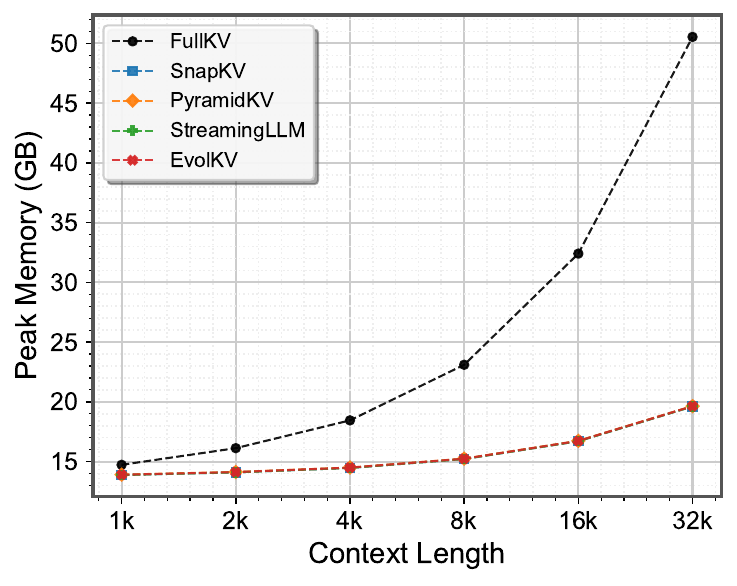}
    \caption{Comparison results of peak memory usage.}
    \label{fig:sub2 peak_memory_res_compare}
  \end{subfigure}
  \caption{Comparison of inference time and peak memory usage between baseline methods and EvolKV.}
  \label{fig:decoding_and_peak_memory_res}
\end{figure*}

We compare the inference time and peak memory usage of EvolKV against baseline methods on Mistral-7B-Instruct, using the NIAH context materials and FlashAttention~\cite{dao2022flashattentionfastmemoryefficientexact} with a target average KV cache budget $c=128$. As shown in Figure~\ref{fig:sub1 decoding_res_compare}, compared to other KV cache compression methods, EvolKV exhibits negligible variation in inference time (prefill + decoding) across different generation lengths. Additionally, as shown in Figure~\ref{fig:sub2 peak_memory_res_compare}, the peak memory usage of EvolKV across different context lengths is comparable to that of other compression methods, while significantly reducing memory consumption compared to the full cache.

\end{document}